\documentclass[11 pt]{article}
\usepackage{style}

\title{\LARGE\bfseries Non-Asymptotic Convergence of Stochastic Iterative Algorithms: A Lyapunov Framework}
\author{Zaiwei Chen\textsuperscript{$*$} and Siva Theja Maguluri\textsuperscript{$\dagger$}\\ 
\small
\textsuperscript{$*$}\textit{Purdue IE,} \href{mailto:chen5252@purdue.edu}{\textit{chen5252@purdue.edu}}\\
\small \textsuperscript{$\dagger$}\textit{Georgia Tech IE,} \href{mailto:siva.theja@gatech.edu}{\textit{siva.theja@gatech.edu}}
}

\date{\vspace{-0.4 in}}
\begin{document}
\maketitle 

\allowdisplaybreaks

\begin{abstract}
We survey Lyapunov-based techniques for the finite-time analysis of stochastic iterative algorithms, also known as stochastic approximation (SA) algorithms, for solving fixed-point equations $\bar{F}(x)=x$, where the operator $\bar{F}(\cdot)$ can only be accessed through a noisy oracle. We first focus on the standard setting in which $\bar{F}(\cdot)$ is contractive with respect to some norm and the noise is i.i.d., and explain how generalized Moreau envelopes serve as universal Lyapunov functions, regardless of the underlying norm. We then show how this framework yields mean-square convergence guarantees and applies to stochastic gradient descent, linear SA, and value-based reinforcement learning algorithms such as Q-learning and temporal-difference learning. Finally, we discuss extensions to Markovian noise, seminorm-contractive operators, dissipative operators, and high-probability bounds, and conclude with open problems. The goal is to present a unified and self-contained roadmap for the finite-time analysis of SA and its applications, especially in reinforcement learning.
\end{abstract}

\section{Introduction}
Root-finding problems, in particular the task of solving fixed-point equations, provide a unifying framework for formulating and analyzing a broad spectrum of computational and analytical tasks that arise across the mathematical sciences. 
For example, minimizing a convex objective function is equivalent to finding the zeros of its gradient operator \citep{beck2017first}. 
In reinforcement learning (RL) \citep{sutton2018reinforcement,bertsekas1996neuro}, which has seen a surge of interest in recent years \citep{silver2016mastering,ouyang2022training}, computing an optimal policy reduces to solving a fixed-point equation known as the Bellman equation \citep{bellman1957dynamic}. 
Additional applications of fixed-point equations arise in game theory and differential equations \citep{border1985fixed,xu2013fixed}.

Formally, let $\bar{F}:\mathbb{R}^d \to \mathbb{R}^d$ be an operator. The fixed-point equation of $\bar{F}(\cdot)$ takes the form $\bar{F}(x) = x$, and any solution is called a fixed point of $\bar{F}(\cdot)$. A natural question is: when does $\bar{F}(x) = x$ admit a (unique) solution? Classical results such as the Banach fixed-point theorem \citep{banach1922operations} provide sufficient conditions. In particular, it states that if $\bar{F}(\cdot)$ is a contraction mapping with respect to some norm, denoted by $\|\cdot\|_c$, then $\bar{F}(x) = x$ has a unique solution $x^*$. Moreover, the fixed-point iteration $x_{k+1}=\bar{F}(x_k)$ converges to $x^*$ at a geometric rate in $\|\cdot\|_c$.

While such fixed-point iterations are appealing in theory, in practice one often lacks the information or computational resources required to evaluate $\bar{F}(\cdot)$ exactly. 
For instance, in large-scale optimization, computing the full gradient may be prohibitively expensive; in RL, the underlying stochastic model is typically unknown, preventing direct evaluation of the Bellman operator. 
In these situations, one must design algorithms that rely on noisy estimates of $\bar{F}(\cdot)$, which leads naturally to the \emph{stochastic approximation} (SA) method~\citep{robbins1951stochastic}. SA forms a foundational framework underlying many modern large-scale methods, including stochastic gradient descent~\citep{bottou2018optimization}, temporal-difference (TD) learning~\citep{sutton1988learning}, Q-learning~\citep{watkins1992q}, and fictitious play \citep{hofbauer2002global}.

The generic SA scheme for solving the fixed-point equation $\bar{F}(x) = x$ takes the form
\begin{align}\label{algo:sa1}    
    x_{k+1} = x_k + \alpha_k \big( F(x_k, Y_k) - x_k \big),\;\forall\, k \ge 0,
\end{align}
where $\{Y_k\}$ is a stochastic process taking values in a set $\mathcal{Y}$, 
$F : \mathbb{R}^d \times \mathcal{Y} \to \mathbb{R}^d$ is an operator, and $\alpha_k > 0$ is the stepsize. 
For simplicity, we assume that $\mathcal{Y}$ is finite, that $\{Y_k\}$ is an i.i.d.\ sequence with distribution $\mu \in \Delta(\mathcal{Y})$,\footnote{Given a finite set $\mathcal{X}$, we use $\Delta(\mathcal{X})$ to denote the set of probability distributions supported on~$\mathcal{X}$.} 
and that the operator $F(\cdot,\cdot)$ satisfies $\mathbb{E}_{Y \sim \mu}[F(x,Y)] = \bar{F}(x)$, 
so that $F(x_k, Y_k)$ provides a conditionally unbiased estimate of $\bar{F}(x_k)$. In later sections, we will discuss several extensions, including settings where the noise process $\{Y_k\}$ is not i.i.d.\ but instead forms a Markov chain, and settings where $\bar{F}(\cdot)$ is not a contraction but admits other useful structural properties.

Early work on SA focused on \emph{asymptotic} convergence. 
A powerful analytical tool in this setting is the \emph{ordinary differential equation} (ODE) method~\citep{borkar2000ode,borkar2009stochastic,kushner2012stochastic,ljung1977analysis}, 
which relates the discrete-time stochastic process defined by the SA iteration~\eqref{algo:sa1} to the continuous-time ODE
\begin{align}\label{eq:ODE}
   \dot{x}(t) = \bar{F}(x(t)) - x(t).
\end{align}
Under mild conditions on the operators $F(\cdot,\cdot)$ and $\bar{F}(\cdot)$, as well as on the decay rate of the stepsize sequence $\{\alpha_k\}$, 
if the equilibrium point $x^*$ of ODE~\eqref{eq:ODE} is globally asymptotically stable~\citep{haddad2011nonlinear,khalil2002nonlinear}, then the SA iterates satisfy $\lim_{k\rightarrow\infty}x_k=x^*$ almost surely~\citep{borkar2009stochastic}.

Although such asymptotic theory is powerful, it does not quantify the \emph{rate} of convergence. 
For many practical applications, particularly in large-scale learning systems, \emph{finite-time} or \emph{non-asymptotic} guarantees are more relevant. 
These guarantees characterize how quickly an algorithm approaches the desired limit and, in data-driven settings, provide concrete guidance on the number of samples or iterations required to achieve a prescribed accuracy \citep{vapnik1999overview}. 
In finite-time analysis, the objective is to understand how the error $\|x_k - x^*\|_c$ decays with $k$. 
Two common performance metrics are the mean squared error and high-probability bounds. 
This survey emphasizes mean squared error while also reviewing results that provide high-probability error guarantees.

A central tool for the finite-time analysis we review is the \emph{Lyapunov-drift approach}. 
Originally developed to study stability in ODEs~\citep{khalil2002nonlinear,haddad2011nonlinear,la2012stability}, 
this method constructs a scalar function $M:\mathbb{R}^d\to\mathbb{R}$ (a Lyapunov or potential function) whose evolution tracks the decay of the ODE towards equilibrium. 
For example, consider the ODE $\dot{x}(t) = -x(t)$, which is a special case of~\eqref{eq:ODE} with $\bar{F}(x) \equiv 0$. For this ODE, the choice $M(x) = \|x\|_2^2$ yields $\dot{M}(x(t)) = -2 M(x(t))$, implying the geometric decay $M(x(t)) = M(x(0)) e^{-2t}$ for any $t \ge 0$. 
By adapting such Lyapunov arguments to the discrete and stochastic setting, one expects to obtain convergence rates for the SA iteration~\eqref{algo:sa1}. However, constructing a valid Lyapunov function is well known to be challenging, and this difficulty is amplified in the context of SA algorithms. 
The Lyapunov function must, on the one hand, induce a negative drift, and on the other hand, enable control of errors arising from stochasticity and discretization. 
In this survey, we present a detailed roadmap for the systematic construction of valid Lyapunov functions tailored to SA.

The remainder of this survey is organized as follows. Section~\ref{sec:sa_mean-square} presents the full mean-square analysis of the SA recursion~\eqref{algo:sa1}, highlighting the construction of a valid Lyapunov function. Section~\ref{sec:RL} discusses applications in RL. Section~\ref{sec:other_sa} reviews several generalizations, such as SA with Markovian noise, SA with seminorm-contractive operators, and SA with dissipative operators, as well as their applications in RL. Section~\ref{sec:sa_concentration} surveys high-probability bounds. We conclude in Section~\ref{sec:future} with open research directions.

\section{Stochastic Approximation: Mean-Square Bounds}\label{sec:sa_mean-square}
Consider the SA recursion (\ref{algo:sa1}). We begin by stating the assumptions regarding the operators $\bar{F}(\cdot)$, $F(\cdot,\cdot)$,    and the noise process $\{Y_k\}$. 

\begin{assumption}\label{as:contraction}
There exist a  norm $\|\cdot\|_c$ and a constant $\gamma_c\in (0,1)$ such that $\|\bar{F}(x_1)-\bar{F}(x_2)\|_c\leq \gamma_c\|x_1-x_2\|_c$ for any $x_1,x_2\in\mathbb{R}^d$.
\end{assumption}

\begin{assumption}\label{as:iid}
The stochastic process $\{Y_k\}$ is an i.i.d. sequence of random variables with distribution $\mu\in\Delta(\mathcal{Y})$.
\end{assumption}
\begin{assumption}\label{as:Lipschitz}
    It holds that $\mathbb{E}_{Y\sim\mu}[F(x,Y)] = \bar{F}(x)$. Moreover, there exist $L_1, L_2 > 0$ such that 
    \begin{enumerate}[(1)]
        \item $\|F(x_1,y) - F(x_2,y)\|_c \leq L_1 \|x_1 - x_2\|_c$ for any $x_1, x_2 \in \mathbb{R}^d$ and $y \in \mathcal{Y}$;
        \item $\|F(0,y)\|_c \leq L_2$ for any $y \in \mathcal{Y}$.
    \end{enumerate}
\end{assumption}

Under Assumption~\ref{as:contraction}, the fixed-point equation $\bar{F}(x)=x$ admits a unique solution, denoted by $x^*$~\citep{banach1922operations}. As a remark, Assumption~\ref{as:contraction} can be relaxed to the setting where $\bar{F}(\cdot)$ is a pseudo-contraction mapping, meaning that $\|\bar{F}(x)-x^*\|_c \leq \gamma_c \|x-x^*\|_c$ for all $x \in \mathbb{R}^d$~\citep{bertsekas1996neuro}. In that case, however, the existence of $x^*$ does not follow automatically and must be imposed separately. Since our goal is to review the Lyapunov approach for analyzing SA algorithms, we adopt Assumption~\ref{as:iid} for ease of presentation. Extensions to more general noise models will be discussed in Section~\ref{subsec:extension_Markov}. Assumption~\ref{as:Lipschitz} is standard even in the classical literature on asymptotic convergence of SA~\citep{borkar2009stochastic}, and it is satisfied in a wide range of applications, such as TD-learning and Q-learning in RL.

In the remainder of this section, we present our detailed approach for establishing the mean-square bounds of $\{x_k\}$ generated by the SA recursion~\eqref{algo:sa1}. Adopting the Lyapunov-drift approach, we begin by illustrating the main challenges in constructing a valid Lyapunov function in Section~\ref{subsec:challenges}. Building on these insights, we then describe a systematic method for constructing a smooth Lyapunov function using the generalized Moreau envelope in Section~\ref{subsubsec:Lyapunov}. With a valid Lyapunov function in hand, we proceed to derive a one-step Lyapunov-drift inequality for the SA recursion in Section~\ref{subsubsec:Drift}, and finally obtain the finite-time bound by solving the resulting recursion in Section~\ref{subsubsec:Recursion}.

\subsection{Fundamental Challenges}\label{subsec:challenges} 
To provide intuition, assume for now that the contraction norm $\|\cdot\|_c$ is the $\ell_p$-norm for some $p \in [2,\infty)$. Consider the ODE~\eqref{eq:ODE} associated with the SA algorithm. It was shown in~\cite[Chapter~10]{borkar2009stochastic} that $W(x) = \|x - x^*\|_p$ satisfies $\dot{W}(x(t)) \le -\kappa W(x(t))$ for some $\kappa > 0$. This inequality implies that the solution $x(t)$ of ODE~\eqref{eq:ODE} converges to its equilibrium point $x^*$ at a geometric rate, which in turn ensures the asymptotic convergence of the SA algorithm via the ODE approach~\citep{ljung1977analysis,borkar2009stochastic}.

The ODE approach yields asymptotic convergence but does not provide finite-time guarantees. To obtain finite-time bounds, we analyze the SA algorithm directly rather than its ODE counterpart. Observe that the SA algorithm in recursion~\eqref{algo:sa1} can be equivalently written as
\begin{align}\label{eq:sa3}
    x_{k+1} - x_k 
    &= \underbrace{\alpha_k \left( \bar{F}(x_k) - x_k \right)}_{\text{Noiseless Update}}
    + \underbrace{\alpha_k \left( F(x_k, Y_k) - \bar{F}(x_k) \right)}_{\text{Effective Noise}}.
\end{align}
Due to discretization and stochastic errors, the ODE Lyapunov function $W(x)$ generally cannot be directly used to analyze the SA algorithm. Suppose there exists a function $M(x)$ that induces a negative drift for the ODE in the sense that $\dot{M}(x(t)) \leq -\kappa' M(x(t))$ and, in addition, is $L$-smooth with respect to some norm, denoted by $\|\cdot\|_s$. Then we can handle both discretization and stochastic errors to obtain
\begin{align}\label{eq:contraction}
	\mathbb{E}[M(x_{k+1} - x^*)] 
	\le \left( 1 - \mathcal{O}(\alpha_k) + o(\alpha_k) \right) \mathbb{E}[M(x_k - x^*)] + o(\alpha_k),
\end{align}
which implies an almost contraction in $\mathbb{E}[M(x_{k+1} - x^*)]$. A finite-time bound then follows by recursively applying~\eqref{eq:contraction}. The key point is that the smoothness of $M(\cdot)$, together with its negative drift in the ODE dynamics, yields a contraction factor of the form $\left( 1 - \mathcal{O}(\alpha_k) + o(\alpha_k) \right)$ for $\{x_k\}$. In the case of $\ell_p$-norm contraction, a natural choice is the norm-square function $M(x) = \|x - x^*\|_p^2$, which is known to be smooth~\citep[Example 5.11]{beck2017first}.
 
In the case where the contraction norm $\|\cdot\|_c$ is arbitrary, the norm-square function $f(x) = \|x - x^*\|_c^2$ is not necessarily smooth, and the key difficulty lies in constructing a smooth Lyapunov function that also induces a negative drift. An important special case is when $\|\cdot\|_c = \|\cdot\|_\infty$, which is particularly relevant in RL, since the Bellman operator is a contraction mapping with respect to $\|\cdot\|_\infty$.

We address this challenge by constructing a smoothed convex envelope $M(x)$, referred to as the \emph{generalized Moreau envelope}, that is smooth and closely approximates the original norm-square function $f(x)$ in the sense that $a M(x) \le f(x) \le b M(x)$ for some constants $a, b$ close to one. The approximation property ensures that $M(\cdot)$ serves as a valid Lyapunov function with a negative drift, while the smoothness property enables us to control both the discretization error and the stochastic error in the SA algorithm. 

\subsection{Construction of the Lyapunov Function}\label{subsubsec:Lyapunov}

To present the construction of a valid Lyapunov function, we need the following definitions.

\begin{definition}\label{df:smooth}
	Let $g:\mathbb{R}^d \to \mathbb{R}$ be a convex and differentiable function. We say that $g(\cdot)$ is $L$-smooth with respect to some norm $\|\cdot\|$ (which can be arbitrary, rather than the $\ell_2$-norm) if
\begin{align*}
    g(y) \le g(x) + \langle \nabla g(x),\, y - x \rangle + \frac{L}{2} \|x - y\|^2, \;\forall\, x,y \in \mathbb{R}^d.
\end{align*}
\end{definition}

Intuitively, the $L$-smoothness ensures that the function does not curve too sharply: all higher-order terms in the Taylor expansion of $g(y)$ around $x$ can be uniformly bounded by a constant multiple of the quadratic term $\|x-y\|^2$. 

Next, we introduce the generalized Moreau envelope.

\begin{definition}\label{df:moreau}
	Let $h_1:\mathbb{R}^d \to \mathbb{R}$ be a closed and convex function, and let $h_2:\mathbb{R}^d \to \mathbb{R}$ be a convex and $L$-smooth function with respect to some norm $\|\cdot\|$. For any $\theta > 0$, the \emph{generalized Moreau envelope} of $h_1(\cdot)$ with respect to $h_2(\cdot)$ is defined as
	\begin{align*}
	    M_{h_1}^{\theta,h_2}(x) = \min_{u \in \mathbb{R}^d} \left\{ h_1(u) + \frac{1}{\theta} h_2(x - u) \right\}.
	\end{align*}
\end{definition}

In the existing literature, $M_{h_1}^{\theta,h_2}(\cdot)$ in Definition~\ref{df:moreau} is also referred to as the \emph{infimal convolution} \citep{beck2017first} between $h_1(\cdot)$ and $h_2(\cdot)/\theta$, and is sometimes denoted by $h_1 \square (h_2/\theta)$ \citep{guzman2015lower,beck2017first}. In the special case where $h_2(x) = \|x\|_2^2$ and $\theta = 1$, $M_{h_1}^{\theta,h_2}(\cdot)$ reduces to the classical \emph{Moreau envelope} \citep{beck2017first}.

Let $f(x) = \frac{1}{2}\|x\|_c^2$, where $\|\cdot\|_c$ is the contraction norm of $\bar{F}(\cdot)$. Let $\|\cdot\|_s$ be an arbitrary norm on $\mathbb{R}^d$ such that $g(x) := \frac{1}{2}\|x\|_s^2$ is $L$-smooth with respect to $\|\cdot\|_s$ for some $L > 0$. For example, $\|\cdot\|_s$ may be chosen as the $\ell_p$-norm for any $p \in [2,\infty)$~\cite[Example 5.11]{beck2017first}. We use the generalized Moreau envelope of $f(\cdot)$ with respect to $g(\cdot)$, i.e.,
\begin{align*}
    M_f^{\theta,g}(x)
    = \min_{u \in \mathbb{R}^d} \left\{ f(u) + \frac{1}{\theta} g(x-u) \right\},
\end{align*}
as our Lyapunov function for analyzing the SA recursion~\eqref{algo:sa1}, where $\theta>0$ is a tunable parameter to be specified. The following proposition shows that $M_f^{\theta,g}(\cdot)$ serves as a \emph{smooth approximation} of the norm-square function $f(\cdot)$. The proof of Proposition~\ref{prop:Moreau} is provided in Appendix~\ref{pf:prop:Moreau}.

\begin{proposition}\label{prop:Moreau}
	The function $M_f^{\theta,g}(\cdot)$ has the following properties.
	\begin{enumerate}[(1)]
	    \item There exists a norm $\|\cdot\|_m$ such that $M_f^{\theta,g}(x) = \frac{1}{2}\|x\|_m^2$.
	    
	    \item Let $\ell_{cs} \in (0,1]$ and $u_{cs} \in [1,\infty)$ be such that 
	    $\ell_{cs}\|\cdot\|_s \le \|\cdot\|_c \le u_{cs}\|\cdot\|_s$. 
	    Then we have $\ell_{cm}\|x\|_m \le \|x\|_c \le u_{cm}\|x\|_m$ for all $x \in \mathbb{R}^d$, 
	    where $\ell_{cm} = (1 + \theta \ell_{cs}^2)^{1/2}$ and $u_{cm} = (1 + \theta u_{cs}^2)^{1/2}$.
	    
        \item $M_f^{\theta,g}(\cdot)$ is convex and $\frac{L}{\theta}$-smooth with respect to $\|\cdot\|_s$.
	\end{enumerate}
\end{proposition}

Proposition~\ref{prop:Moreau} (1) states that this generalized Moreau envelope is itself a norm-square function. 
Propositions~\ref{prop:Moreau} (2) and (3) together show that $M_f^{\theta,g}(\cdot)$ provides a smooth approximation of $f(\cdot)$.

\subsection{Establishing a One-Step Contractive Inequality}\label{subsubsec:Drift}

For simplicity of notation, we henceforth write $M(x)$ for $M_f^{\theta,g}(x)$. For any $k \ge 0$, by the smoothness of $M(\cdot)$ and the SA recursion~\eqref{algo:sa1}, we have
\begin{align}\label{eq:composition1_contractive}
    M(x_{k+1} - x^*)
    \leq \;& M(x_k - x^*) + \langle \nabla M(x_k - x^*), x_{k+1} - x_k \rangle+ \frac{L}{2\theta} \|x_{k+1} - x_k\|_s^2 \nonumber\\
    =\; & M(x_k - x^*) + \alpha_k \langle \nabla M(x_k - x^*), F(x_k, Y_k) - x_k \rangle \nonumber\\
    &+ \frac{L \alpha_k^2}{2\theta} \|F(x_k, Y_k) - x_k\|_s^2 \nonumber\\
    =\; & M(x_k - x^*) 
    + \underbrace{\alpha_k \langle \nabla M(x_k - x^*), \bar{F}(x_k) - x_k \rangle}_{E_1:\ \text{The expected update}} \nonumber\\
    &+ \underbrace{\alpha_k \langle \nabla M(x_k - x^*), F(x_k, Y_k) - \bar{F}(x_k) \rangle}_{E_2:\ \text{The error due to noise } Y_k} \nonumber\\
    &+ \underbrace{\frac{L \alpha_k^2}{2\theta} \|F(x_k, Y_k) - x_k\|_s^2}_{E_3:\ \text{The error due to discretization and noise}}.
\end{align}

The term $E_1$ on the right-hand side of Inequality~\eqref{eq:composition1_contractive} captures the deterministic part of the SA iteration~\eqref{algo:sa1}, and is bounded in the following lemma. The proof of Lemma~\ref{le:T1} is provided in Appendix~\ref{pf:le:T1}.

\begin{lemma}\label{le:T1}
    The following inequality holds for all $k \ge 0$: 
    \begin{align*}
        E_1 \leq - 2\alpha_k \left(1 - \frac{\gamma_c u_{cm}}{\ell_{cm}}\right) M(x_k - x^*).
    \end{align*}
\end{lemma}
By choosing the tunable parameter $\theta$ such that $1 - \gamma_c\, u_{cm} / \ell_{cm} \in (0,1)$, which is always feasible since $\lim_{\theta \to 0} u_{cm} / \ell_{cm} = 1$ (cf.\ Proposition~\ref{prop:Moreau}) and $\gamma_c \in (0,1)$, Lemma~\ref{le:T1} ensures a negative drift.

We next consider the term $E_2$ in Inequality~\eqref{eq:composition1_contractive}. Since $\{Y_k\}$ is an i.i.d.\ sequence, we have by the tower property of conditional expectations that
\begin{align}
    \mathbb{E}[E_2] 
    = \alpha_k \mathbb{E} \left[ \left\langle \nabla M(x_k - x^*),\, \mathbb{E}[F(x_k, Y_k) \mid \mathcal{F}_k] - \bar{F}(x_k) \right\rangle \right] = 0, \label{eq:T_2}
\end{align}
where $\mathcal{F}_k$ denotes the $\sigma$-algebra generated by $\{Y_n\}_{0 \le n \le k-1}$. 
Equation~\eqref{eq:T_2} follows from the fact that $x_k$ is measurable with respect to $\mathcal{F}_k$ and $Y_k$ is independent of $\mathcal{F}_k$. 
In the more general setting where $\{Y_k\}$ forms a Markov chain, more sophisticated arguments are required to control $\mathbb{E}[E_2]$, which will be discussed in Section~\ref{subsec:extension_Markov}.

Lastly, we bound the term $E_3$ from Inequality~\eqref{eq:composition1_contractive} in the following lemma. The proof of Lemma~\ref{le:T3} relies primarily on the Lipschitz continuity of $F(\cdot,\cdot)$ (cf.\ Assumption~\ref{as:Lipschitz}) and is deferred to Appendix~\ref{pf:le:T3}.

\begin{lemma}\label{le:T3}
    It holds for any $k \ge 0$ that 
    \begin{align*}
        E_3\leq  \frac{2L(L_1+1)^2u_{cm}^2\alpha_k^2}{\theta\ell_{cs}^2}M(x_k-x^*)+\frac{L\alpha_k^2}{\theta\ell_{cs}^2}\left((L_1+1)\|x^*\|_c+L_2\right)^2
    \end{align*}
\end{lemma}

Using Lemma~\ref{le:T1}, Equation~\eqref{eq:T_2}, and Lemma~\ref{le:T3} together in Inequality~\eqref{eq:composition1_contractive}, we obtain the following one-step contractive inequality, whose proof is presented in Appendix~\ref{pf:le:MSE_drift}. For simplicity of notation, we denote $M_k = \mathbb{E}[M(x_k - x^*)]$.

\begin{lemma}\label{le:MSE_drift}
    When $\alpha_0\leq \varphi_0:=\frac{\theta \ell_{cs}^2\left(1 - \gamma_c u_{cm}/\ell_{cm}\right)}{2 L (L_1+1)^2 u_{cm}^2 }$, the following inequality holds for all $k \geq 0$:
    \begin{align}\label{eq:recursion}
        M_{k+1} \leq \left(1-\left(1 - \frac{\gamma_c u_{cm}}{\ell_{cm}}\right)\alpha_k \right)M_k+\frac{L\alpha_k^2}{\theta\ell_{cs}^2}\left((L_1+1)\|x^*\|_c+L_2\right)^2.
    \end{align}   
\end{lemma}

Lemma~\ref{le:MSE_drift} has the desired form~\eqref{eq:contraction}, showing that the SA recursion~\eqref{algo:sa1} is overall contractive with respect to the Lyapunov function $M(\cdot)$ up to higher-order error terms.

\subsection{Solving the Recursion}\label{subsubsec:Recursion}
The last step is to solve the recursion~\eqref{eq:recursion} and evaluate the resulting bounds under stepsizes with different decay rates. Let $\varphi_2 = 1 - \gamma_c u_{cm} / \ell_{cm}$. By repeatedly applying Inequality~\eqref{eq:recursion}, we have for all $k \ge 0$ that
\begin{align*}
    M_k
    \leq \prod_{j=0}^{k-1}\left(1-\varphi_2\alpha_j\right)M_0
    + \frac{L}{\theta \ell_{cs}^2} \big( (L_1+1)\|x^*\|_c + L_2 \big)^2
    \sum_{i=0}^{k-1}\alpha_i^2 \prod_{j=i+1}^{k-1}\left(1-\varphi_2\alpha_j\right).
\end{align*}
After using Proposition~\ref{prop:Moreau} (2) to translate $M_k$ back to the mean squared error $\mathbb{E}[\|x_k-x^*\|_c^2]$, we obtain for all $k \ge 0$ that
\begin{align}\label{eq:before_solving}
    \mathbb{E}[\|x_k - x^*\|_c^2]
    \leq\;& \varphi_1 \prod_{j=0}^{k-1}\left(1-\varphi_2\alpha_j\right)
    + \varphi_3 \sum_{i=0}^{k-1}\alpha_i^2 \prod_{j=i+1}^{k-1}\left(1-\varphi_2\alpha_j\right),
\end{align}
where
\begin{align*}
    \varphi_1 = \frac{\|x_0 - x^*\|_c^2 u_{cm}^2}{\ell_{cm}^2},
    \qquad
    \varphi_3 = \frac{2 u_{cm}^2 L}{\theta \ell_{cs}^2} \big( (L_1+1)\|x^*\|_c + L_2 \big)^2.
\end{align*}
The final step is to evaluate the right-hand side of Inequality~\eqref{eq:before_solving} for different choices of stepsizes. Specifically, we consider stepsizes of the form $\alpha_k = \alpha/(k+h)^z$, where $\alpha, h > 0$ and $z \in [0,1]$. Note that $z = 0$ corresponds to a constant stepsize, while $z \in (0,1]$ corresponds to diminishing stepsizes. Evaluating such terms under this type of stepsize schedule is well studied in the convergence-rate analysis of iterative algorithms~\citep{srikant2019finite,beck2017first} and is omitted here; see~\citep[Appendix~A.8]{chen2024lyapunov} for additional details.

\subsection{Finite-Time Mean-Square Bounds}
After evaluating Inequality~\eqref{eq:before_solving} under the specified stepsize rule, we obtain the finite-time bounds for the SA recursion~\eqref{algo:sa1} stated in the following theorem.

\begin{theorem}\label{thm:sa_MSE}
Suppose that Assumptions~\ref{as:contraction}--\ref{as:Lipschitz} hold, and that $\alpha_k=\alpha/(k+h)^z$, where $z\in [0,1]$ and $\alpha,h$ are chosen such that $\alpha_0 \le \varphi_0$. Then we have the following results.
\begin{enumerate}[(1)]
\item When $z=0$, for all $k \ge 0$,
\begin{align*}
    \mathbb{E}[\|x_k - x^*\|_c^2]
    \le \varphi_1 (1 - \varphi_2 \alpha)^k + \frac{\varphi_3}{\varphi_2}\,\alpha.
\end{align*}

\item When $z=1$, for all $k \ge 0$,
\begin{align*}
    \mathbb{E}[\|x_k - x^*\|_c^2]
    \le 
    \begin{dcases}
        \varphi_1 \left(\frac{h}{k+h}\right)^{\varphi_2 \alpha}
        + \frac{8 \alpha^2 \varphi_3}{1 - \varphi_2 \alpha}\frac{1}{(k+h)^{\varphi_2 \alpha}},
        & \alpha < 1/\varphi_2, \\begin{align*}0.35em]
        \varphi_1 \frac{h}{k+h}
        + \frac{8 \alpha^2 \varphi_3 \log(k+h)}{k+h},
        & \alpha = 1/\varphi_2, \\begin{align*}0.35em]
        \varphi_1 \left(\frac{h}{k+h}\right)^{\varphi_2 \alpha}
        + \frac{4 e \varphi_3 \alpha^2}{\varphi_2 \alpha - 1}\frac{1}{k+h},
        & \alpha > 1/\varphi_2.
    \end{dcases}
\end{align*}

\item When $z\in (0,1)$, for all $k \ge 0$,
\begin{align*}
    \mathbb{E}[\|x_k - x^*\|_c^2]
    \le \varphi_1
    e^{-\frac{\varphi_2 \alpha}{1-z} \left( (k+h)^{1-z} - (h)^{1-z} \right)}
    + \frac{4 \varphi_3 \alpha}{\varphi_2}\frac{1}{(k+h)^z}.
\end{align*}
\end{enumerate}

\end{theorem}

In view of Theorem~\ref{thm:sa_MSE}, it is clear that the choice of stepsizes plays an important role in determining the convergence rate of the SA algorithm. Before illustrating each case of Theorem~\ref{thm:sa_MSE} in detail, we provide a high-level intuition. For ease of illustration, consider the special case where the contraction norm $\|\cdot\|_c$ is the Euclidean norm $\|\cdot\|_2$. In this case, we have the following standard decomposition:
\begin{align*}
    \underbrace{\mathbb{E}[\|x_k - x^*\|_2^2]}_{\text{Mean Squared Error}}
    =
    \underbrace{\|\mathbb{E}[x_k] - x^*\|_2^2}_{\text{Squared bias}}
    +
    \underbrace{\mathbb{E}[\|x_k - \mathbb{E}[x_k]\|_2^2]}_{\text{Variance}}.
\end{align*}

Suppose there is no noise, i.e., $F(\cdot, y) \equiv \bar{F}(\cdot)$ for any $y \in \mathcal{Y}$, so the variance term vanishes and the mean-square error reduces to the squared bias. In this case, we should choose a large stepsize to maximize the per-step improvement. This is indeed the case: choosing $\alpha_k \equiv 1$ recovers the fixed-point iteration and yields geometric convergence. When there is noise, however, a trade-off arises in choosing the stepsizes. The bias term favors large stepsizes, as discussed above, while the variance term favors small stepsizes. This is because at each iteration the effective noise injected into $x_{k+1}$ is $\alpha_k (F(x_k, Y_k) - \bar{F}(x_k))$, whose variance is proportional to $\alpha_k^2$. With this high-level intuition regarding the bias–variance trade-off in choosing the stepsizes in mind, we next discuss the three cases presented in Theorem~\ref{thm:sa_MSE}.

In all cases of Theorem~\ref{thm:sa_MSE}, the convergence bound consists of a bias term (the first term on the right-hand side of the inequality) and a variance term (the second term on the right-hand side). A constant stepsize yields geometric decay of the bias with a constant variance that is proportional to the stepsize. When using diminishing stepsizes of the form $\alpha_k = \alpha/(k+h)$, and when the parameter $\alpha$ is above a certain threshold, namely $1/\varphi_2$, both terms decay to zero at a rate of $\mathcal{O}(1/k)$. However, when $\alpha$ is not chosen properly, the convergence rate can be arbitrarily bad. We will return to this issue in Section~\ref{sec:future} when discussing open problems in SA. Finally, when using diminishing stepsizes of the form $\alpha_k = \alpha/(k+h)^z$ with $z \in (0,1)$, the bias term decays to zero at an almost geometric rate, while the variance term decays at a suboptimal rate of $\mathcal{O}(1/k^z)$, so the overall convergence rate is  $\mathcal{O}(1/k^z)$. Importantly, the convergence rate in this case, while being sub-optimal, is independent of $\alpha$ (which only appears as a multiplicative constant), and is therefore more robust.

\subsection{Connection to SGD and linear SA}
In this section, we illustrate that while we present the finite-time mean squared bound for a generic contractive SA framework, the result also captures several other popular SA settings, including SGD and linear SA.

\paragraph{Stochastic Gradient Descent.} Consider the problem $\min_{x\in\mathbb{R}^d} J(x)$. With access to a noisy oracle that returns a stochastic estimate of the gradient $\nabla J(x)$, the SGD updates $x_k$ according to
\begin{align*}
    x_{k+1} = x_k + \beta_k \big( -\nabla J(x_k) + w_k \big),
\end{align*}
where $w_k$ denotes the mean-zero noise and $\beta_k$ is the stepsize. For any $\eta>0$, since the SGD update rule can be equivalently written as
\begin{align*}
    x_{k+1}
    = x_k + \frac{\beta_k}{\eta} \left( -\eta \nabla J(x_k) + x_k + \eta w_k - x_k \right),
\end{align*}
by defining $\alpha_k = \beta_k / \eta$, $Y_k = \eta w_k$, and $F(x,y) = -\eta \nabla J(x) + x + y$, the SGD update can be formulated as an instance of the SA recursion~\eqref{algo:sa1}. Moreover, under the assumption that $J(\cdot)$ is both smooth and strongly convex, with parameters $L_J$ and $\sigma_J$, respectively, the operator $\bar{F}(x) = -\eta \nabla J(x) + x$ is Lipschitz with respect to the Euclidean norm $\|\cdot\|_2$, with Lipschitz constant $L_{\mathrm{SGD}} = \max(|1 - \eta \sigma_J|,\, |1 - \eta L_J|)$~\citep{ryu2016primer}. Therefore, when $\eta \in (0, 2/L_J)$, we have $L_{\mathrm{SGD}} < 1$, and hence the operator $\bar{F}(\cdot)$ is a contraction with respect to $\|\cdot\|_2$. As a result, Theorem~\ref{thm:sa_MSE} applies to SGD in this setting.

\paragraph{Linear Stochastic Approximation.} A linear SA is a recursion of the form
\begin{align}\label{eq:linear_SA}
    x_{k+1} = x_k + \beta_k (A_k x_k - b_k),
\end{align}
where $\{A_k\}$ and $\{b_k\}$ are sequences of i.i.d.\ random matrices and vectors, respectively, and $\beta_k$ is the stepsize. Letting $\bar{A} = \mathbb{E}[A_k]$ and $\bar{b} = \mathbb{E}[b_k]$, the linear SA can be viewed as iteratively solving the linear system of equations $\bar{A}x = \bar{b}$. Popular algorithms such as linear regression and a large class of TD-learning-based algorithms in RL, e.g., TD$(0)$, $n$-step TD, TD$(\lambda)$, off-policy TD-learning, and TD-learning with linear function approximation, can all be formulated as linear SA of the form presented above.

Assuming that $\bar{A}$ is Hurwitz, that is, all of its eigenvalues have negative real parts, such a linear SA algorithm can be reformulated as a contractive SA in the form of~\eqref{algo:sa1}. Specifically, for any $\eta > 0$, observe that the update rule~\eqref{eq:linear_SA} is equivalent to
\begin{align*}
    x_{k+1}
    = x_k + \frac{\beta_k}{\eta} \big( (I + \eta A_k)x_k - \eta b_k - x_k \big).
\end{align*}
Therefore, by defining $\alpha_k = \beta_k / \eta$, $Y_k = (A_k, b_k)$, and $F(x,y)=F(x,A,b) = (I + \eta A)x - \eta b$, the linear SA presented above is a special instance of the generic SA recursion~\eqref{algo:sa1}. Moreover, since $\bar{A}$ is Hurwitz, there exists a unique positive definite matrix $P$ that solves the Lyapunov equation $\bar{A}^\top P + P \bar{A} + I = 0$ \citep{khalil2002nonlinear}.
As a result, with an appropriately chosen $\eta$, the operator $\bar{F}(x) = (I + \eta \bar{A})x - \eta \bar{b}$ is a contraction mapping with respect to the weighted norm $\|x\|_P = \sqrt{x^\top P x}$. To see this, note that for any $x_1,x_2\in\mathbb{R}^d$, we have
\begin{align*}
    \|\bar{F}(x_1) - \bar{F}(x_2)\|_P^2
    =\,& (x_1 - x_2)^\top (I + \eta \bar{A})^\top P (I + \eta \bar{A})(x_1 - x_2)\\
    =\,& (x_1 - x_2)^\top \big( P + \eta \bar{A}^\top P + \eta P \bar{A} + \eta^2 \bar{A}^\top P \bar{A} \big)(x_1 - x_2)\\
    =\,& (x_1 - x_2)^\top \big( P - \eta I + \eta^2 \bar{A}^\top P \bar{A} \big)(x_1 - x_2)\\
    \le\,&\left( 1 - \frac{\eta}{\lambda_{\max}(P)}
    + \frac{\eta^2 \lambda_{\max}(\bar{A}^\top P \bar{A})}{\lambda_{\min}(P)} \right)
    \|x_1 - x_2\|_P^2\\
    =\,&\left( 1 - \frac{\lambda_{\min}(P)}
        {4 \lambda_{\max}(\bar{A}^\top P \bar{A}) \lambda_{\max}(P)^2} \right)
        \|x_1 - x_2\|_P^2,
\end{align*}
where the last inequality follows by choosing
\begin{align*}
    \eta = \frac{\lambda_{\min}(P)}
        {2 \lambda_{\max}(\bar{A}^\top P \bar{A}) \lambda_{\max}(P)}.
\end{align*}
Since the operator $\bar{F}(\cdot)$ is a contraction with respect to $\|\cdot\|_P$, Theorem~\ref{sec:sa_mean-square} applies to linear SA.

\subsection{Related Work}
The SA method was originally proposed in \cite{robbins1951stochastic} for solving root-finding problems under a noisy oracle. 

\paragraph{The Asymptotic Results.} 
The asymptotic convergence of SA has been studied extensively in the literature. 
A central line of work is based on the ODE method, which compares the interpolated 
trajectory of the discrete-time recursion with the flow of the mean ODE. Under 
standard stability and noise assumptions, if the stepsizes are non-summable and 
square summable, and if the relevant equilibrium point of the ODE is globally 
asymptotically stable, then the SA iterates converge almost surely 
\citep{benveniste2012adaptive,kushner2012stochastic,borkar2009stochastic,ljung1977analysis}. 
Such almost-sure convergence results are quite general: the ODE can be replaced by 
a differential inclusion, which is useful for set-valued or nonsmooth mean fields 
\citep{benaim2005stochastic,benaim2006stochastic}, and the classical square-summability 
condition on the stepsizes can be relaxed in several directions 
\citep{lauand2024revisiting,nguyen2026almost}. 

Beyond almost-sure convergence, a parallel line of work studies asymptotic rates and 
distributional limits. For smooth mean fields, after linearizing the mean drift 
$\bar F(x)-x$ around a locally stable equilibrium $x^*$, the suitably normalized 
error process satisfies central-limit-theorem-type results; for instance, under 
stepsizes of order $1/k$, the last iterate typically has $\sqrt{k}$-scale fluctuations 
whose covariance is determined by the Jacobian of the mean field and the covariance 
of the martingale noise 
\citep{chung1954stochastic,sacks1958asymptotic,fabian1968asymptotic,benveniste2012adaptive}. 
These results provide asymptotic convergence rates and form the basis for statistical 
inference based on SA iterates. A major refinement is Polyak--Ruppert averaging, 
which shows that averaging the iterates can achieve the optimal asymptotic covariance 
under substantially less delicate stepsize tuning than is required for the last iterate 
\citep{ruppert1988efficient,polyak1992acceleration}. More recent CLT and asymptotic 
efficiency results have further extended this theory to controlled Markovian dynamics, 
parameter-dependent Markovian noise, nonsmooth problems, and decision-dependent data 
distributions 
\citep{fort2015central,borkar2025ode,davis2024asymptotic,cutler2024stochastic}.

\paragraph{The Non-Asymptotic Results.} 
Finite-time analysis has received a lot of attention in the last decade 
\citep{lan2020first,moulines2011non,srikant2019finite}. A central approach for 
establishing mean-square bounds is the Lyapunov method: one constructs a scalar 
function that simultaneously captures the stability of the limiting dynamics and 
permits control of the stochastic and discretization errors. In the strongly stable 
case, this typically leads to a one-step recursion consisting of a contracting bias 
term and a variance term. For example, with a constant stepsize, such recursions 
often yield a geometrically decaying transient plus an $O(\alpha)$ steady-state 
error, whereas with a properly tuned diminishing stepsize they yield an $O(1/k)$ 
mean-square rate.

For linear SA \citep{srikant2019finite,lakshminarayanan2018linear,mou2020linear}, 
the Lyapunov function is usually constructed from the solution of a Lyapunov 
equation associated with the Hurwitz mean matrix 
\citep{haddad2011nonlinear,khalil2002nonlinear}. This quadratic structure has led 
to sharp finite-time mean-square bounds for both last iterates and averaged 
iterates, including results that clarify the role of constant stepsizes and 
Polyak--Ruppert averaging. For nonlinear SA, finite-time analysis is more delicate, 
and existing results typically exploit additional structure, such as stochastic 
gradient methods and their variants 
\citep{lan2020first,moulines2011non,duchi2012ergodic,doan2022finite,bansal2019potential}, 
contractive SA \citep{wainwright2019stochastic,qu2020finite,chen2020finite,chen2024lyapunov}, 
or SA under dissipativity assumptions \citep{chen2022automatica}. These structural 
conditions play the role of producing a negative drift. Beyond standard mean-square bounds, finite-time analysis can be carried out in a more refined way by separating 
the optimization bias, the asymptotic bias, and the variance 
\citep{zhang2024prelimit,zhang2024constant}.

Another important direction concerns weakly stable operators, especially 
nonexpansive operators. The study of SA under nonexpansive maps has a long history 
in reinforcement learning \citep{abounadi2002stochastic}, and recent work has 
developed finite-time analyses for stochastic approximation with non-expansive operators. In 
this setting, because strict contraction and uniqueness of the fixed point may fail, 
the guarantees are often stated in terms of fixed-point residuals rather than 
mean-square distance to a distinguished fixed point 
\citep{bravo2024stochastic}. More recently, finite-sample analyses have also been 
developed for nonexpansive SA with Markovian noise  
\citep{blaser2026asymptotic}. 

\section{Applications in Reinforcement Learning}\label{sec:RL}
In this section, we show how to use the SA results presented in the previous section to establish finite-time bounds of popular RL algorithms.
RL provides a principled framework for sequential decision-making under uncertainty \citep{sutton2018reinforcement}, with broad applications in game playing \citep{silver2017mastering}, robotics \cite{levine2016end}, recommendation systems \citep{afsar2022reinforcement}, and large language models (LLMs) \citep{ouyang2022training}, inspiring a surge of theoretical research aimed at deepening the mathematical foundations of RL and guiding its practical deployment. 

More formally, an RL problem is usually modeled as a Markov decision process (MDP)~\citep{puterman2014markov}. In this survey, we consider a discounted infinite-horizon MDP consisting of a finite state space $\mathcal{S}$, a finite action space $\mathcal{A}$, a transition kernel $\{p(\cdot \mid s,a)\}_{(s,a)\in\mathcal{S}\times \mathcal{A}}$, a reward function $\mathcal{R}:\mathcal{S}\times\mathcal{A}\to[0,1]$, and a discount factor $\gamma\in(0,1)$. Since we work with finite MDPs, assuming bounded rewards is indeed without loss of generality. Notably, the parameters of the stochastic model, such as the reward function and the transition kernel, are unknown to the agent. 

Given a stationary policy $\pi:\mathcal{S}\to\Delta(\mathcal{A})$, the associated value function $V^\pi:\mathcal{S}\to\mathbb{R}$ and Q-function $Q^\pi:\mathcal{S}\times\mathcal{A}\to\mathbb{R}$ are defined as 
\begin{align*}
    V^\pi(s)=\,&\mathbb{E}_\pi\!\left[\sum_{k=0}^\infty\gamma^k\mathcal{R}(S_k,A_k)\,\middle|\, S_0=s\right],\;\forall\,s\in\mathcal{S},\\
    Q^\pi(s,a)=\,&\mathbb{E}_\pi\!\left[\sum_{k=0}^\infty\gamma^k\mathcal{R}(S_k,A_k)\,\middle|\, S_0=s, A_0=a\right],\;\forall\,(s,a)\in\mathcal{S}\times \mathcal{A},
\end{align*}
respectively, 
where we use $\mathbb{E}_\pi[\cdot]$ to indicate that the actions are selected according to the policy $\pi$. Alternatively, the value function and the Q-function can be viewed as vectors in $\mathbb{R}^{|\mathcal{S}|}$ and $\mathbb{R}^{|\mathcal{S}||\mathcal{A}|}$.

The goal is to find an optimal policy $\pi^*$ so that its value function $V^*$, or equivalently its Q-function $Q^*$, is uniformly maximized~\citep{sutton2018reinforcement,bertsekas1996neuro}; that is, $V^*(s)\ge V^\pi(s)$ for all $\pi$ and $s$, respectively, $Q^*(s,a)\ge Q^\pi(s,a)$ for all $\pi$ and $(s,a)$. It is known that such an optimal policy always exists~\citep{bertsekas1996neuro}. 

In RL, the problem of finding an optimal policy is called the \textit{control} problem. One of the most popular algorithms for solving the control problem is Q-learning~\citep{watkins1992q}. While the ultimate goal is to find an optimal policy, there is usually a smaller goal of finding the value function of a given policy, which is called the \textit{prediction} problem and is typically solved with TD-learning and its variants~\citep{sutton1988learning}. Both Q-learning and TD-learning are, in nature, SA algorithms for solving their associated Bellman equations. Therefore, the results of SA unify the finite-time analysis of value-based RL algorithms. We next present a detailed case study using Q-learning.

\subsection{Q-Learning}
The Q-learning algorithm proposed in~\cite{watkins1992q} finds an optimal policy $\pi^*$ by computing the optimal Q-function $Q^* = Q^{\pi^*}$. The motivation for Q-learning is based on the following fact: $\pi$ is an optimal policy if and only if $\{a \mid \pi(a\mid s)>0\} \subseteq \arg\max_{a\in\mathcal{A}} Q^*(s,a)$ for all $s\in\mathcal{S}$~\citep{sutton2018reinforcement,bertsekas1996neuro}. Therefore, finding the optimal Q-function is sufficient for finding an optimal policy.

We next present the Bellman equation for the optimal Q-function $Q^*$, which serves as the foundation for the design of Q-learning. Let $\mathcal{H}:\mathbb{R}^{|\mathcal{S}||\mathcal{A}|}\to\mathbb{R}^{|\mathcal{S}||\mathcal{A}|}$ be the Bellman operator defined as 
\begin{align*}
    [\mathcal{H}(Q)](s,a)
    = \mathcal{R}(s,a)
    + \gamma\,\mathbb{E}\!\left[\max_{a'\in\mathcal{A}} Q(S_{k+1},a')
    \;\middle|\; S_k=s, A_k=a\right]
\end{align*}
for all $Q\in\mathbb{R}^{|\mathcal{S}||\mathcal{A}|}$ and $(s,a)$. Then, the Bellman equation is simply the fixed-point equation of the Bellman operator $\mathcal{H}(\cdot)$:
\begin{align*}
    Q = \mathcal{H}(Q).
\end{align*}
It is well known that $\mathcal{H}(\cdot)$ is a contraction mapping with respect to $\|\cdot\|_\infty$, with $Q^*$ as its unique fixed point~\citep{puterman2014markov}. Therefore, a natural approach to finding $Q^*$ is to perform the fixed-point iteration $Q_{k+1} = \mathcal{H}(Q_k)$. However, carrying out such a fixed-point iteration requires complete knowledge of the transition probabilities of the underlying MDP, and hence is not feasible in RL when the stochastic model is unknown. The Q-learning algorithm is then proposed as a data-driven stochastic variant of this fixed-point iteration.

Let $\pi_b$ be the behavior policy, i.e., the policy used to collect samples. Assuming that the Markov chain $\{S_k\}$ induced by $\pi_b$ is irreducible, it admits a unique stationary distribution $\kappa_b \in \Delta(\mathcal{S})$ satisfying $\kappa_b(s) > 0$ for all $s \in \mathcal{S}$ \citep{levin2017markov}. Then, with a sequence of i.i.d.\ samples $\{(S_k, A_k, S_k')\}$ generated as $S_k \sim \kappa_b(\cdot)$, $A_k \sim \pi_b(\cdot \mid S_k)$, and $S_k' \sim p(\cdot \mid S_k, A_k)$ for all $k \ge 0$, the Q-learning algorithm is given in Algorithm~\ref{algo:Q-learning}. In this survey paper, we assume i.i.d.\ sampling for ease of exposition. Finite-time bounds for Q-learning under Markovian sampling can be established using results on Markovian SA, to be presented in later sections (see Theorem~\ref{thm:Markovian_sa}).

\begin{algorithm}[ht]\caption{Q-Learning with i.i.d.\ Sampling}\label{algo:Q-learning}
	\begin{algorithmic}[1]
		\STATE \textbf{Input:} Initialization $Q_0$, and a sample trajectory $\{(S_k,A_k,S_k')\}_{k\ge 0}$.
		\FOR{$k=0,1,2,\cdots$}
        \STATE Compute the temporal difference $\delta_k=\mathcal{R}(S_k,A_k)+\gamma \max_{a'} Q_k(S_k',a')-Q_k(S_k,A_k)$.
		\STATE Update the Q-function according to 
        \begin{align*}
            Q_{k+1}(s,a)
            = Q_k(s,a) + \alpha_k \mathds{1}_{\{(S_k,A_k)=(s,a)\}} \delta_k,\quad \forall\,(s,a)\in\mathcal{S}\times\mathcal{A}.
        \end{align*}
		\ENDFOR
	\end{algorithmic}
\end{algorithm}

To establish the finite-time bounds for Q-learning, we first reformulate Q-learning in the form of the SA recursion (\ref{algo:sa1}). Let $Y_k=(S_k,A_k,S_k')\in\mathcal{Y}:=\mathcal{S}\times\mathcal{A}\times\mathcal{S}$ for all $k\ge 0$. It is clear that $\{Y_k\}$ is an i.i.d.\ sequence with distribution $\mu\in\Delta(\mathcal{Y})$ satisfying $\mu(s,a,s')=\kappa_b(s)\pi_b(a\mid s)p(s'\mid s,a)$. Let $F:\mathbb{R}^{|\mathcal{S}||\mathcal{A}|}\times\mathcal{Y}\to\mathbb{R}^{|\mathcal{S}||\mathcal{A}|}$ be an operator defined such that, given inputs $Q\in\mathbb{R}^{|\mathcal{S}||\mathcal{A}|}$ and $y_0=(s_0,a_0,s_1)$, the $(s,a)$-th component of the output is 
\begin{align*}
	[F(Q,y_0)](s,a)
    = \begin{dcases}
        \mathcal{R}(s_0,a_0)+\gamma\max_{a'\in\mathcal{A}}Q(s_1,a'),&(s,a)=(s_0,a_0),\\
        Q(s,a),&(s,a)\neq (s_0,a_0).\\
    \end{dcases}
\end{align*}
Then, the update equation of the Q-learning algorithm presented in Line $4$ of Algorithm~\ref{algo:Q-learning} can be equivalently written as
\begin{align*}
	Q_{k+1} = Q_k + \alpha_k \left(F(Q_k,Y_k)-Q_k\right),\;\forall\,k\geq 0,
\end{align*}
which is in the same form as the SA recursion (\ref{algo:sa1}).

The next step is to verify that all assumptions required to apply Theorem~\ref{thm:sa_MSE} are satisfied in the context of Q-learning. Assumption~\ref{as:iid} is clearly satisfied since $\{Y_k\}$ is a sequence of i.i.d.\ samples. We next show that Assumptions~\ref{as:contraction} and~\ref{as:Lipschitz} are satisfied. To present the result, let $D_b$ be an $|\mathcal{S}||\mathcal{A}| \times |\mathcal{S}||\mathcal{A}|$ diagonal matrix with diagonal components $\{\kappa_b(s)\pi_b(a \mid s)\}_{(s,a)\in\mathcal{S}\times \mathcal{A}}$. Denote the minimum diagonal entry of $D_b$ by $D_{b,\min}$. The proof of the following proposition can be found in~\cite{chen2024lyapunov}.

\begin{proposition}\label{prop:Q-learning}
	The operator $\bar{F}(\cdot):=\mathbb{E}_{Y\sim \mu}[F(\cdot,Y)]$ is explicitly given as $\bar{F}(Q)=D_b\mathcal{H}(Q)+(I-D_b)Q$ for all $Q\in\mathbb{R}^{|\mathcal{S}||\mathcal{A}|}$.
	In addition, we have the following results.
    \begin{enumerate}[(1)]
        \item $\bar{F}(\cdot)$ is a $\hat{\gamma}_c$-contraction mapping with respect to $\|\cdot\|_\infty$, where $\hat{\gamma}_c=1-D_{b,\min}(1-\gamma)$;
        \item It holds for all $Q_1,Q_2$ and $y=(s,a,s')$ that $\|F(Q_1,y)-F(Q_2,y)\|_\infty\leq 2\|Q_1-Q_2\|_\infty$ and $\|F(0,y)\|_\infty\leq 1$.
    \end{enumerate}
\end{proposition}

We next present the mean-square bounds for Q-learning. For simplicity, we focus on the case of constant stepsize; the extension to diminishing stepsizes is straightforward in view of Theorem \ref{thm:sa_MSE}. 

\begin{theorem}\label{thm:Q}
	Suppose that $\alpha_k \equiv \alpha$ is appropriately chosen. Then for all $k \ge 0$,
	\begin{align*}
		\mathbb{E}[\|Q_k - Q^*\|_\infty^2]
		\le\;& c_1\|Q_0 - Q^*\|_\infty^2
        \left(1 - \frac{D_{b,\min}(1-\gamma)\alpha}{2}\right)^{k} +\frac{c_2\log(|\mathcal{S}||\mathcal{A}|)}{D_{b,\min}^2(1-\gamma)^2}
        (\|Q^*\|_\infty + 1)^2 \alpha,
	\end{align*}
    where $c_1,c_2$ are absolute constants.
	As a result, given $\epsilon>0$, to achieve $\mathbb{E}[\|Q_k - Q^*\|_\infty] \le \epsilon$, the sample complexity is $\tilde{\mathcal{O}}(\epsilon^{-2}(1-\gamma)^{-5}D_{b,\min}^{-3})$.
\end{theorem}

\begin{remark}
Since the SA framework presented here is a general tool and is not specifically 
designed for Q-learning, which admits other properties such as uniform boundedness 
of the iterates \citep{gosavi2006boundedness}, the sample complexity of Q-learning 
presented in this work does not match the statistical lower bound 
\citep{gheshlaghi2013minimax}. To achieve the minimax lower bound, other advanced 
algorithmic ideas (such as variance reduction) and a more tailored analysis are 
needed; see \cite{li2020sample} for more details.
\end{remark}

Beyond Q-learning, most value-based RL algorithms can be modeled as SA algorithms under contractive operators. Consequently, the SA results presented in this survey paper apply to a broad class of RL methods. We briefly discuss a few examples below.

\subsection{The Efficiency of Bootstrapping in TD-Learning} 
Given a target policy $\pi$, consider the problem of predicting its value function $V^\pi$ (or its Q-function $Q^\pi$). Such problems are typically solved with TD-learning \citep{sutton1988learning} and its variants. At a high level, TD-learning estimates $V^\pi$ iteratively using a sample trajectory collected by applying the target policy $\pi$ to interact with the environment. An important idea for potentially improving the performance of TD-learning is to use bootstrapping, with typical examples including $n$-step TD and TD$(\lambda)$~\citep{sutton2018reinforcement}. Intuitively, the degree of bootstrapping is determined by how much the current estimate depends on previous estimates. For instance, in $n$-step TD, $n=1$ corresponds to extreme bootstrapping (the next iterate depends solely on the previous one), whereas $n=\infty$ corresponds to no bootstrapping at all, in which case the method reduces to Monte Carlo.

Despite strong empirical evidence, theoretically analyzing the effect of bootstrapping remains challenging. It is listed as one of the four major open problems in~\cite{sutton1999open}, which states that ``\textit{While it remains unclear exactly what should or could be proved here, it is clear that this [referring to the efficiency of bootstrapping] is a key open question at the heart of current and future RL.}'' The results presented here for general contractive SA algorithms allow us to establish finite-time bounds for $n$-step TD (and TD$(\lambda)$) as explicit functions of $n$ (and $\lambda$), thereby providing theoretical insight into the efficiency of bootstrapping. We emphasize that the open problem is not resolved here: to provide a definitive answer, upper bounds alone are insufficient; a matching lower bound is needed.

As an aside, since TD-learning is often used within an actor--critic framework~\citep{konda2000actor} to ultimately compute an optimal policy, the finite-time analyses of TD-learning also enable a deeper understanding of the performance of a variety of actor--critic algorithms.

\subsection{The Bias-Variance Trade-Off in Off-Policy TD-Learning}
Given a target policy $\pi$, the TD-learning algorithm for predicting $V^\pi$ (or $Q^\pi$) can be divided into two categories: on-policy TD and off-policy TD. In on-policy TD-learning, the samples are collected under the target policy $\pi$, which was discussed in the previous section. In off-policy TD, samples are generated by a \textit{behavior} policy $\pi_b \neq \pi$. Off-policy sampling is used for three important reasons: (1) it is typically necessary to incorporate exploration in the behavior policy $\pi_b$, which makes it differ from the target policy $\pi$; (2) it is widely used in multi-agent training, where different agents collect rewards using behavior policies that lag behind the target policies in an actor--critic framework \cite{espeholt2018impala}; and (3) it enables learning from historical data, thereby improving sample efficiency.

In practice, off-policy TD-learning is implemented via importance sampling to obtain an unbiased estimate of the desired value function. However, the product of importance sampling ratios induces high variance in the estimate \citep{glynn1989importance}, which is a fundamental issue in off-policy learning. To mitigate this problem, many variants of off-policy TD-learning have been proposed, including the $Q^\pi(\lambda)$ algorithm \citep{harutyunyan2016q}, the $\mathrm{TB}(\lambda)$ algorithm \citep{precup2000eligibility}, the $\mathrm{Retrace}(\lambda)$ algorithm \citep{munos2016safe}, and the Q-trace algorithm \citep{chen2024lyapunov}. These methods modify the importance sampling ratios in various ways to reduce variance, albeit at the cost of introducing bias in the limit. While numerical experiments can compare their empirical performance, the lack of analytical tools has made the bias--variance trade-off in tuning the importance sampling ratios unclear.

The results for generic contractive SA presented in this survey paper enable a unified theoretical analysis of all these algorithms and provide quantitative characterizations of the associated bias--variance trade-offs; see~\cite{chen2021off} for further details.

\section{Beyond Contractive Stochastic Approximation with i.i.d.\ Noise}\label{sec:other_sa}
So far, we have focused on SA under a contractive operator with i.i.d. noise sequences. In this section, we discuss several extensions of the SA framework.

\subsection{Stochastic Approximation under Markovian Noise}\label{subsec:extension_Markov}

In RL, the samples, namely trajectories of state-action pairs, are typically collected by an agent implementing a behavior policy to interact with the environment. In this case, the resulting sample trajectory forms a Markov chain rather than an i.i.d.\ sequence~\citep{bertsekas1996neuro}. In this section, we discuss how to extend the finite-time analysis to the Markovian noise setting.

To begin, we state the assumption that formalizes the notion of Markovian noise.

\begin{assumption}\label{as:Markovian}
	$\{Y_k\}$ is a finite, irreducible, and aperiodic Markov chain.
\end{assumption}

\begin{remark}
    In general, aperiodicity is not a necessary requirement for a Markovian SA to converge. We will later present two approaches to relax the aperiodicity assumption.
\end{remark}

Assumption \ref{as:Markovian} implies that $\{Y_k\}$ has a unique stationary distribution $\mu\in\Delta(\mathcal{Y})$. In this case, the expected operator $\bar{F}(\cdot):=\mathbb{E}_{Y\sim \mu}[F(\cdot,Y)]$ is defined with respect to the stationary distribution $\mu$. In addition, $\{Y_k\}$ mixes at a geometric rate; that is, there exist $C>0$ and $\rho\in(0,1)$ such that $\max_{y\in\mathcal{Y}}\|P^k(y,\cdot)-\mu(\cdot)\|_{\text{TV}} \le C\rho^k$ for all $k\ge 0$, where $\|\cdot\|_{\text{TV}}$ denotes the total variation distance and $P\in\mathbb{R}^{|\mathcal{Y}|\times|\mathcal{Y}|}$ is the transition probability matrix of $\{Y_k\}$ \citep{levin2017markov}. Let $t_\alpha := \min\left\{k \ge 0 : \max_{y \in \mathcal{Y}} \|P^{k}(y,\cdot) - \mu(\cdot)\|_{\text{TV}} \le \alpha\right\}$, which captures how fast the distribution of $Y_k$ converges to its stationary distribution $\mu$ and is called the mixing time of the Markov chain $\{Y_k\}$ with precision $\alpha$. Note that under Assumption \ref{as:Markovian}, we have $t_\alpha = \mathcal{O}(\log(1/\alpha))$.

We next state the finite-time bounds for Markovian SA under contractive operators.

\begin{theorem}\label{thm:Markovian_sa}
	Consider $\{x_k\}$ generated by the SA recursion (\ref{algo:sa1}). Suppose that Assumptions \ref{as:contraction}, \ref{as:Lipschitz}, and \ref{as:Markovian} are satisfied, and that $\alpha_k \equiv \alpha$ is small enough. Then we have:
\begin{align}\label{eq:bound:Markovian}
    \mathbb{E}[\|x_k - x^*\|_c^2]
    \le \bar{\varphi}_1 (1 - \bar{\varphi}_2 \alpha)^{k - t_\alpha} + \bar{\varphi}_3 \alpha t_\alpha,\quad \forall\,k\geq t_\alpha,
\end{align}
where $\bar{\varphi}_1, \bar{\varphi}_3 > 0$ and $\bar{\varphi}_2 \in (0,1)$ are problem-dependent constants.
\end{theorem}

For ease of exposition, we consider only constant stepsizes in Theorem \ref{thm:Markovian_sa}. See \cite{chen2024lyapunov} for results under diminishing stepsizes.

Compared with the finite-time bound of SA under i.i.d.\ noise (cf.\ Theorem \ref{thm:sa_MSE} (1)), there are two differences in Theorem \ref{thm:Markovian_sa}. The first is that the bound holds only when $k \ge t_\alpha$ instead of for all $t \ge 0$. Intuitively, this is because the distribution of $Y_k$ is not ``stable'' until $\{Y_k\}$ is sufficiently mixed. Moreover, the second term on the right-hand side of Inequality (\ref{eq:bound:Markovian}) includes an additional factor of $t_\alpha = \mathcal{O}(\log(1/\alpha))$, which also arises from the Markovian noise. Other than these two points, Theorem \ref{thm:Markovian_sa} is qualitatively similar to Theorem \ref{thm:sa_MSE}, as the mean squared error consists of a geometrically decaying bias and a constant variance of order $\mathcal{O}(\alpha \log(1/\alpha))$.

To establish Theorem~\ref{thm:Markovian_sa}, the high-level idea is identical to that of Theorem~\ref{thm:sa_MSE} in Section~\ref{sec:sa_mean-square}, namely to use the generalized Moreau envelope as a Lyapunov function and to establish a one-step contractive inequality. After the error decomposition in Inequality~\eqref{eq:composition1_contractive}, the main challenge under Markovian noise is to control the expected value of the term $E_2$ from Inequality~\eqref{eq:composition1_contractive}, which we rewrite below as $\hat{E}_2$ to avoid confusion:
\begin{align*}
    \hat{E}_2 = \alpha_k \mathbb{E}[\langle \nabla M(x_k - x^*),\, F(x_k, Y_k) - \bar{F}(x_k) \rangle].
\end{align*}
In view of Lemma \ref{le:T1}, since the negative drift is $\mathcal{O}(\alpha_k)$, we need to show that $\hat{E}_2 = o(\alpha_k)$ so that it is dominated by the negative drift. When $\{Y_k\}$ is an i.i.d.\ sequence, as considered in Section \ref{sec:sa_mean-square}, we immediately have $\hat{E}_2 = 0$ by the tower property of conditional expectations. When $\{Y_k\}$ is a Markov chain, controlling $\hat{E}_2$ requires additional effort.

We next present the high-level idea of an approach for bounding $\hat{E}_2$ using a conditioning argument, which was first developed in \cite{srikant2019finite} for linear SA and later extended in \cite{chen2022automatica} to nonlinear SA. The first step is to rewrite the term $\hat{E}_2$ as
\begin{align}\label{eq:markov_decomposition}
    \hat{E}_2
    = \alpha_k \mathbb{E}[\langle \nabla M(x_{k-t} - x^*),\, F(x_{k-t}, Y_k) - \bar{F}(x_{k-t}) \rangle]
    + \textit{residual terms},
\end{align}
where $t$ is a positive integer yet to be chosen, and the residual terms arise from replacing $x_k$ with $x_{k-t}$ from the original expression of $\hat{E}_2$.

For the first term on the right-hand side of (\ref{eq:markov_decomposition}), by applying the tower property of conditional expectations, we have
\begin{align*}
    &\alpha_k \mathbb{E}[\langle \nabla M(x_{k-t} - x^*),\, F(x_{k-t}, Y_k) - \bar{F}(x_{k-t}) \rangle] \\
    =\;& \alpha_k \mathbb{E}[\langle \nabla M(x_{k-t} - x^*),\, \mathbb{E}[F(x_{k-t}, Y_k) \mid \mathcal{F}_{k-t}] - \bar{F}(x_{k-t}) \rangle] \\
    \leq \;& \alpha_k \mathcal{O}(C \rho^t),
\end{align*}
where $\mathcal{F}_{k-t}$ is the $\sigma$-algebra generated by $\{Y_n\}_{0 \le n \le k-t-1}$, and the last inequality can be established using the geometric mixing of $\{Y_k\}$. As for the \textit{residual terms} in (\ref{eq:markov_decomposition}), since there are $t$ updates required to obtain $x_k$ from $x_{k-t}$, and each update contributes an amount proportional to the stepsize, we can show that
\begin{align*}
    \textit{residual terms}
    = \mathcal{O}\!\left(\alpha_k \sum_{\ell = k-t}^k \alpha_\ell \right).
\end{align*}
Finally, by choosing $t = \mathcal{O}(\log(1/\alpha_k))$, we obtain $\hat{E}_2 = o(\alpha_k)$ as desired. See \cite{srikant2019finite,chen2022automatica} for more details regarding the conditioning argument for dealing with Markovian noise.

\paragraph{Relaxing the Aperiodicity Assumption.} To relax the aperiodicity requirement, the most straightforward approach is to apply an aperiodic transformation to the data samples. Specifically, for a finite Markov chain that is irreducible, the corresponding lazy chain is both irreducible and aperiodic. Moreover, with an aperiodicity transformation, also known as the Schweitzer transformation~\citep{schweitzer1971iterative}, one can construct a trajectory from the original sample path as if it were generated by the lazy transition matrix. 

Specifically, let $\{Y_k\}$ be a finite and irreducible Markov chain with transition matrix $P$. Fix some $\eta\in(0,1)$ and define the lazy transition matrix $\tilde{P}=(1-\eta)P+\eta I$. Given a realized trajectory $\{Y_k\}$, we construct a new trajectory $\{Y_k'\}$ as follows. Let $Y_0'=Y_0$. For each $k\ge 0$, insert a geometrically distributed number of copies of $Y_k$ (with success probability $1-\eta$) before moving to $Y_{k+1}$. In other words, between $Y_k$ and $Y_{k+1}$ in the original trajectory, we repeat the state $Y_k$ a random number of times, and then append $Y_{k+1}$. The resulting process $\{Y_k'\}$ is a Markov chain with transition matrix $\tilde{P}$. Since $P$ is irreducible and $\eta>0$, the matrix $\tilde{P}$ is irreducible and aperiodic. Therefore, by working with the modified trajectory $\{Y_k'\}$ constructed from the original sample path, one may assume without loss of generality that the underlying Markov chain is aperiodic.

While the aforementioned approach is theoretically sound, it is practically inconvenient, since one needs to repeatedly insert identical samples to justify the analysis. A more direct but technically more challenging approach is to use the Poisson equation to decompose the Markovian noise. For each fixed $x$, define the forcing function $h_x(y)=F(x,y)-\bar{F}(x)$. The Poisson equation is then given by $\hat{h}_x-P\hat{h}_x=h_x$, where $\hat{h}_x$ is the solution (unique up to an additive constant) \citep{douc2018markov}. Using the Poisson equation, we write $F(x_k,Y_k)-\bar{F}(x_k)=h_{x_k}(Y_k)$ from $\hat{E}_2$ in \eqref{eq:markov_decomposition} as 
\begin{align*}
    F(x_k,Y_k)-\bar{F}(x_k)=h_{x_k}(Y_k)=\hat{h}_{x_k}(Y_k)-P\hat{h}_{x_k}(Y_k).
\end{align*}
Adding and subtracting $\hat{h}_{x_k}(Y_{k+1})$, we obtain
\begin{align*}
    F(x_k,Y_k)-\bar{F}(x_k)
=\big(\hat{h}_{x_k}(Y_k)-\hat{h}_{x_k}(Y_{k+1})\big)
+\big(\hat{h}_{x_k}(Y_{k+1})-P\hat{h}_{x_k}(Y_k)\big).
\end{align*}
The second term satisfies 
$\mathbb{E}[\hat{h}_{x_k}(Y_{k+1})-P\hat{h}_{x_k}(Y_k)\mid \mathcal{F}_k]=0$, 
so it forms a martingale difference sequence, while the first term telescopes across iterations. This representation separates the Markovian dependence from the stochastic fluctuations, allowing the finite-time analysis to proceed by combining martingale arguments with bounds on the Poisson solution $\hat{h}_x$. A detailed illustration of the Poisson equation-based approach for dealing with Markovian noise can be found in \cite{chandak2021concentration,haque2025stochastic,nanda2025minimal}.

\subsection{Stochastic Approximation under Seminorm Contractive Operators}\label{subsec:seminorm_SA}

In this section, instead of imposing the contraction assumption, we consider SA under seminorm contractive operators. This setting is motivated by average-reward RL, where the Bellman operator is not contractive in any norm but can be seminorm contractive under certain structural assumptions on the underlying MDP \citep{puterman2014markov}, and by control problems in which only partial stability can be established \citep{haddad2011nonlinear}. We begin by defining what we mean by a seminorm.

\begin{definition}
	\label{definition:seminorm}
	A real-valued function $p:\mathbb{R}^d \rightarrow \mathbb{R}$ is called a seminorm if and only if it is nonnegative and satisfies: (1) \textit{triangle inequality}: $p(x_1 + x_2) \le p(x_1) + p(x_2)$ for all $x_1, x_2 \in \mathbb{R}^d$, and (2) \textit{absolute homogeneity}: $p(\alpha x) = |\alpha|\, p(x)$ for all $x \in \mathbb{R}^d$ and $\alpha \in \mathbb{R}$.
\end{definition}

Recall that a norm $\|\cdot\|$ must also satisfy $\|x\| = 0$ if and only if $x = 0$, a property that is not required for a seminorm. Before proceeding, we present two representative examples of seminorms.
\begin{itemize}
    \item The span seminorm $\textit{sp}(\cdot)$ is defined as $\textit{sp}(x) = \max_i x_i - \min_j x_j$ for all $x \in \mathbb{R}^d$. One can easily verify that $\textit{sp}(\cdot)$ is indeed a seminorm. The span seminorm is widely used in studying the convergence behavior of learning algorithms for solving average-reward MDPs \citep{puterman2014markov}.
    \item Let $P \in \mathbb{R}^{d \times d}$ be a positive semidefinite matrix. Then the function $p(x) = \sqrt{x^\top P x}$ is a seminorm. Such seminorms are widely used in the analysis of least-squares methods and in control, where quadratic Lyapunov functions naturally induce such seminorm structures \citep{haddad2011nonlinear}.
\end{itemize}

Returning to the SA recursion (\ref{algo:sa1}), we impose the following assumptions on the operators $\bar{F}(\cdot)$ and $F(\cdot,\cdot)$.
\begin{assumption}\label{as:semi-norm}
	There exist a seminorm $p_c(\cdot)$ and a constant $\gamma_c\in (0,1)$ such that $p_c(\bar{F}(x_1)-\bar{F}(x_2))\leq \gamma_cp_c(x_1-x_2)$ for all $x_1,x_2\in\mathbb{R}^d$.
\end{assumption}
\begin{assumption}\label{as:seminorm_Lipschitz}
    It holds that $\mathbb{E}_{Y\sim\mu}[F(x,Y)] = \bar{F}(x)$. Moreover, there exist $L_1', L_2' > 0$ such that 
    \begin{enumerate}[(1)]
        \item $p_c(F(x_1,y) - F(x_2,y)) \leq L_1' p_c(x_1 - x_2)$ for any $x_1, x_2 \in \mathbb{R}^d$ and $y \in \mathcal{Y}$;
        \item $p_c(F(0,y)) \leq L_2'$ for any $y \in \mathcal{Y}$.
    \end{enumerate}
\end{assumption}

Similar to the case of norm-contractive operators, seminorm-contractive operators also enjoy several useful properties. In particular, the set of solutions to the seminorm fixed-point equation $p_c(\bar{F}(x) - x) = 0$ must be of the form $x^* + \ker(p_c)$, where $x^*$ is a particular solution and $\ker(p_c) = \{x \mid p_c(x) = 0\}$. Moreover, the fixed-point iteration $x_{k+1} = \bar{F}(x_k)$ converges geometrically in the seminorm: $p_c(x_k - x^*) \le \gamma_c^k p_c(x_0 - x^*)$. See~\cite{chen2025non} for more details. We next state the finite-time bounds of SA under seminorm-contractive operators.

\begin{theorem}\label{thm:sa_seminorm}
	Consider $\{x_k\}$ generated by the SA recursion (\ref{algo:sa1}). Suppose that Assumptions~\ref{as:semi-norm}, \ref{as:seminorm_Lipschitz}, and \ref{as:iid} hold and that $\alpha_k \equiv \alpha$ is chosen sufficiently small. Then, for all $k\geq 0$,
	\begin{align*}
		\mathbb{E}[p_c(x_k-x^*)^2]
		\leq  \hat{\varphi}_1(1-\hat{\varphi}_2\alpha)^{k}+\hat{\varphi}_3\alpha,
	\end{align*}
	where  $\hat{\varphi}_1,\hat{\varphi}_3>0$ and $\hat{\varphi}_2\in (0,1)$ are (problem-dependent) constants.
\end{theorem}

Theorem~\ref{thm:sa_seminorm} is qualitatively similar to Theorem~\ref{thm:sa_MSE}, except that the norm is replaced by the seminorm. The proof of Theorem~\ref{thm:sa_seminorm} also follows a Lyapunov approach, where we choose the following variant of the generalized Moreau envelope,
\begin{align}\label{average_Lyapunov}
    \hat{M}(x) := \min_{u \in \mathbb{R}^d} \left\{ \frac{1}{2} p_c^2(u) + \frac{1}{2\theta} \|x - u\|_s^2 \right\},
\end{align}
as the Lyapunov function. The resulting analysis is similar to that of Theorem~\ref{thm:sa_MSE}; see~\cite{chen2025non} for more details.

\subsection{Stochastic Approximation under a Dissipativity Assumption}\label{subsec:SA_dissipative} 
The following assumption introduces a dissipative operator.

\begin{assumption}\label{as:stability}
	There exists a unique $x^* \in \mathbb{R}^d$ such that $\bar{F}(x^*) = x^*$. Moreover, there exist a positive definite matrix $\bar{P}$ and a constant $\bar{c}_0 > 0$ such that, for all $x \in \mathbb{R}^d$,
\begin{align*}
    (x - x^*)^\top \bar{P}\bigl(\bar{F}(x) - x\bigr) \le -\bar{c}_0 (x - x^*)^\top \bar{P} (x - x^*).
\end{align*}

\end{assumption}

Assumption \ref{as:stability} can be viewed as an exponential dissipativity property of the ODE~\eqref{eq:ODE} with the quadratic storage function $x^\top \bar{P} x$ \citep{haddad2011nonlinear}, which motivates referring to Assumption \ref{as:stability} as the dissipativity assumption. A special case in which Assumption \ref{as:stability} holds is linear SA with a Hurwitz matrix, i.e.,
\begin{align*}
    x_{k+1}
    =\;& x_k + \alpha_k \bigl(A(Y_k)x_k - b(Y_k)\bigr) \\
    =\;& x_k + \alpha_k\bigl(\underbrace{(A(Y_k) + I_d)x_k - b(Y_k)}_{F(x_k,Y_k)} - x_k\bigr),
\end{align*}
where $A:\mathcal{Y} \to \mathbb{R}^{d \times d}$ and $b:\mathcal{Y} \to \mathbb{R}^d$ are deterministic functions. Denote $\bar{A} = \mathbb{E}_{Y \sim \mu}[A(Y)]$, assumed to be Hurwitz, and $\bar{b} = \mathbb{E}_{Y \sim \mu}[b(Y)]$. In this case, $\bar{F}(x) = (\bar{A} + I_d)x - \bar{b}$ and the unique fixed point of $\bar{F}(\cdot)$ is $x^* = \bar{A}^{-1}\bar{b}$. Since $\bar{A}$ is Hurwitz, there exists a unique positive definite matrix $\bar{P}$ such that
\begin{align*}
    \bar{A}^\top \bar{P} + \bar{P}\bar{A} + I_d = 0
\end{align*}
\citep{khalil2002nonlinear}. Consequently, for any $x \in \mathbb{R}^d$,
\begin{align*}
    (x - x^*)^\top \bar{P}\bigl(\bar{F}(x) - x\bigr)
    =\;& (x - x^*)^\top \bar{P}(\bar{A}x - \bar{b}) \\
    =\;& (x - x^*)^\top \bar{P}\bar{A}(x - x^*) \\
    =\;& \tfrac{1}{2}(x - x^*)^\top\bigl(\bar{A}^\top \bar{P} + \bar{P}\bar{A}\bigr)(x - x^*) \\
    =\;& -\tfrac{1}{2}(x - x^*)^\top(x - x^*) \\
    \le\;& -\tfrac{1}{2\lambda_{\max}(\bar{P})} (x - x^*)^\top \bar{P}(x - x^*),
\end{align*}
where $\lambda_{\max}(\bar{P})$ is the largest eigenvalue of $\bar{P}$. Hence, Assumption \ref{as:stability} is satisfied.

Under Assumptions \ref{as:iid}, \ref{as:Lipschitz}, and \ref{as:stability}, using $\tilde{M}(x) = x^\top \bar{P} x$ as the Lyapunov function and following the approach in Section~\ref{sec:sa_mean-square}, we can establish finite-time guarantees for the SA algorithm with update \eqref{algo:sa1}, achieving a qualitatively similar rate of convergence as in Theorem~\ref{thm:sa_MSE}. See \cite{chen2022automatica} for more details.

\subsection{Other Applications in Reinforcement Learning}

The extensions (SA with Markovian noise, semi-norm contractive SA, and SA under a dissipativity assumption) presented in earlier sections enables studying many other RL algorithms. We will briefly talk about a few in the following.

\subsubsection{Q-Learning with Linear Function Approximation} 
In practice, RL algorithms are typically combined with function approximation to overcome the curse of dimensionality, which refers to the computational bottleneck as the size of the state-action space increases. One of the most successful practical algorithms is the deep Q-network, which is essentially Q-learning with neural network approximation \citep{mnih2015human}. On the other hand, the behavior of Q-learning with function approximation is not yet fully understood theoretically and was identified in \citep{sutton1999open} as an open problem. In fact, the infamous deadly triad \citep{sutton2018reinforcement} arises in Q-learning with function approximation, and hence even in the basic setting with linear function approximation, the algorithm can be unstable in general \citep{baird1995residual}. 

The SA results in the previous sections not only allow us to analyze existing RL algorithms but also enable the development of new RL algorithms with theoretical performance guarantees. Specifically, they lead to a variant of Q-learning with linear function approximation that incorporates a target network and truncation, is provably stable, and achieves a $\tilde{\mathcal{O}}(\epsilon^{-2})$ sample complexity up to a function approximation error \citep{chen2022target}.

\subsubsection{Robust Reinforcement Learning}
Robust RL studies sequential decision-making when the transition model used for 
training may differ from the transition model encountered at deployment. This is 
often formulated through robust MDPs, where the goal is to optimize the worst-case 
value over an ambiguity set of transition kernels. In the discounted setting, under 
rectangular uncertainty sets, the robust Bellman operator retains a contraction 
structure. This contraction property is the key reason why value-based robust RL algorithms can be analyzed through the same contractive-SA perspective.

This connection is made explicit in \citep{wang2023finite}. Although the robust Bellman update involves an inner worst-case optimization problem, the induced robust Bellman operator remains contractive in $\|\cdot\|_\infty$. Hence, the analysis uses the generalized Moreau-envelope Lyapunov function to establish the one-step drift bound. The same SA viewpoint is used in robust policy optimization. In 
\citep{li2022first}, stochastic robust policy mirror descent relies on repeated 
robust policy evaluation from samples. The resulting robust TD subroutine is a 
Markovian contractive SA recursion, so Theorem~\ref{thm:Markovian_sa} provides 
the finite-time critic bound, which is then propagated through the mirror-descent 
analysis to obtain the overall sample-complexity guarantee. More recently, \citep{xu2025finite} applied this Lyapunov approach to robust 
average-reward policy evaluation. Since discounting is absent, the key structural step is to show that the robust average-reward Bellman operator is contractive under a suitable seminorm. The robust TD recursion is then analyzed with a Moreau-envelope-type Lyapunov function, analogous to $\hat M(\cdot)$ in Equation~\eqref{average_Lyapunov}.

\subsubsection{Federated Reinforcement Learning}
Federated RL studies settings where multiple agents collaboratively learn a 
shared value function or $Q$-function while keeping their local trajectories, 
data, and possibly behavior policies private. This setting is natural in RL 
because sampling is often the computational bottleneck, and parallel data 
collection across $N$ agents can in principle reduce the sample complexity. 

In \citep{khodadadian2022federated}, the authors extend the Markovian-SA 
Lyapunov framework to this federated setting. They introduce a general 
federated stochastic approximation scheme with Markovian noise, 
which includes federated on-policy TD-learning, off-policy TD-learning, and 
Q-learning as special cases. The analysis again makes use of the generalized 
Moreau-envelope Lyapunov function to obtain a one-step drift inequality for 
the global averaged iterate. 

\section{Stochastic Approximation: High Probability Bounds}\label{sec:sa_concentration}
In this section, we survey results on high-probability bounds for the SA recursion~\eqref{algo:sa1}, i.e., bounds on the quantity $\mathbb{P}(\|x_k - x^*\|_c \le \epsilon)$ for some $\epsilon > 0$. High-probability bounds are often preferable to mean-square bounds because they provide both a convergence rate and a confidence level. A naive way to obtain such bounds from a mean-square guarantee is to apply Markov's inequality \citep{vershynin2018high}; however, this yields only polynomially decaying tails. Establishing high-probability bounds with super-polynomial tails (i.e., tails lighter than polynomial tails) is substantially more challenging. For example, even in the classical setting of the law of large numbers,\footnote{The sample average $\frac{1}{k}\sum_{i=0}^{k-1} Y_i$ can be computed iteratively as $x_{k+1} = x_k + \frac{1}{k+1}(-x_k + Y_k)$ with $x_0 = 0$, which is a special case of Algorithm~\eqref{algo:sa1}.} deriving exponential tail bounds, such as Hoeffding's inequality, the Chernoff bound, or Bernstein's inequality, requires significantly more work.

To establish high-probability bounds with super-polynomial tails for the SA recursion~\eqref{algo:sa1}, the tail behavior of the effective noise term $F(x_k, Y_k) - \bar{F}(x_k)$ plays a central role. Although Assumption~\ref{as:iid} ensures that this noise is conditionally mean zero, its tail behavior depends on the interplay between the random process $\{Y_k\}$ and the iterates $\{x_k\}$, in particular, whether $Y_k$ is additive or multiplicative with respect to $x_k$. In the remainder of this section, we focus on the additive (yet unbounded) noise setting. The Lyapunov technique introduced here can be extended to the multiplicative noise case, but the latter requires substantially more technical work.

\subsection{Stochastic Approximation under Sub-Gaussian Additive Noise}\label{sec:additive}
Consider the SA recursion~\eqref{algo:sa1}. In addition to Assumptions~\ref{as:contraction} and \ref{as:iid}, we impose the following condition on the tail behavior of the effective noise $F(x_k, Y_k) - \bar{F}(x_k)$. Let $\mathcal{F}_k$ denote the $\sigma$-algebra generated by $\{Y_n\}_{0 \le n \le k-1}$.
\begin{assumption}\label{ass:sub-Gaussian}
There exist $\Bar{\sigma} > 0$ and a (possibly) dimension-dependent constant $c_d > 0$ such that 
the following inequalities hold for any $k\geq 0$ and $\mathcal{F}_k$-measurable $d$-dimensional random vector $v$:
\begin{align}
    &\mathbb{E}\!\left[\exp\!\left(\lambda \langle F(x_k,Y_k) \!-\! \bar{F}(x_k), v \rangle \right) \middle| \mathcal{F}_k \right]
    \!\le\! \exp\!\left(\lambda^2 \Bar{\sigma}^2 \|v\|_{c,*}^2 / 2\right), \; \forall\, \lambda > 0, \label{eq:sub-Gaussian_1} \\
    &\mathbb{E}\!\left[\exp\!\left(\lambda \left\|F(x_k,Y_k) \!-\! \bar{F}(x_k)\right\|_c^2 \right) \middle| \mathcal{F}_k \right]
    \le \left(1 \!-\! 2\lambda \Bar{\sigma}^2\right)^{-\frac{c_d}{2}}, \;\forall\, \lambda \in \left(0, \frac{1}{2\Bar{\sigma}^2} \right), \label{eq:sub-Gaussian_2}
\end{align}
where $\|\cdot\|_{c,*}$ is the dual norm of the contraction norm $\|\cdot\|_c$.
\end{assumption}

Assumption~\ref{ass:sub-Gaussian} generalizes the standard notion of a random vector being norm sub-Gaussian \citep{jin2019short} to the setting where an arbitrary norm $\|\cdot\|_c$ is used in place of $\|\cdot\|_2$. Indeed, when $\|\cdot\|_c = \|\cdot\|_2$ and $c_d = d$, Inequalities~\eqref{eq:sub-Gaussian_1} and~\eqref{eq:sub-Gaussian_2} specialize exactly to the equivalent definitions of a sub-Gaussian random vector \citep{jin2019short,wainwright2019high}. Since we allow an arbitrary norm, we correspondingly allow a (possibly different) dimension-dependent constant $c_d$. A representative setting in which Assumptions~\ref{as:iid}, \ref{as:contraction}, and~\ref{ass:sub-Gaussian} all hold is when the noise $Y_k$ is purely additive and forms an i.i.d.\ mean-zero sequence with sub-Gaussian tails.

We next state the high-probability bound of the SA recursion (\ref{algo:sa1}). 

\begin{theorem}[\cite{chen2023concentration}]\label{thm:additive}
Consider $\{x_k\}$ generated by recursion \eqref{algo:sa1}. Suppose Assumptions \ref{as:contraction}, \ref{as:iid}, and \ref{ass:sub-Gaussian} are satisfied and $\alpha_k=\alpha/(k+h)$ with $\alpha$ and $h$ chosen appropriately. Then, for any $\delta > 0$ and $K \ge 0$, we have with probability at least $1-\delta$ that, for all $k \ge K$,
\begin{align*}
    \|x_k - x^*\|_c^2 
    \le\;& \frac{\bar{c}_1 \log(1/\delta)}{k+h}
    + \frac{\bar{c}_2 h\|x_0 - x^*\|_c^2+\bar{c}_3 + \bar{c}_4 \log\bigl((k+1)/K^{1/2}\bigr)}{k+h},
\end{align*}
where $\{\bar{c}_i\}_{1\leq i\leq 4}$ are problem-dependent constants.
\end{theorem}

The first term captures the deterministic contraction of the initial error, while the remaining terms capture the stochastic fluctuations. The logarithmic factors arise from controlling the martingale terms uniformly over time. Next, we discuss the implications of Theorem \ref{thm:additive} in terms of its dependence on $\delta$, $K$, and $k$. Since the tolerance level $\delta$ appears as $\log(1/\delta)$ in the norm-square bound, the norm error $\|x_k - x^*\|_c$ has a sub-Gaussian tail. As for the dependence on $K$ and $k$, Theorem \ref{thm:additive} implies that, with probability at least $1-\delta$, all the iterates lie in a cone with radius $\tilde{\Theta}((\log(1/\delta)^{1/2}+\log(k/K^{1/2}))k^{-1/2})$ for all $k \ge K$. As byproducts of Theorem~\ref{thm:additive}, one can also establish high-probability bounds for a fixed time $K$ (rather than for all $k \ge K$), as well as the full tail bound on the quantity $\mathbb{P}(\|x_k - x^*\|_c > \epsilon)$. Consequently, using the formula
\begin{align*}
    \mathbb{E}[\|x_k - x^*\|_c^r]
    = \int_0^\infty \mathbb{P}(\|x_k - x^*\|_c^r > x)\, dx
\end{align*}
for any positive integer $r$, we can integrate the tail bound to obtain moment bounds for the error at any fixed time. These results are omitted here.

\subsection{High Probability Bounds of Q-Learning}
Theorem \ref{thm:additive} enables us to establish the concentration bounds of Q-learning (cf. Algorithm \ref{algo:Q-learning}) for solving the RL problem, which is presented in the following.
\begin{theorem}[\cite{chen2023concentration}]\label{thm:Q-learning}
Suppose that $\{(S_k,A_k,S_k')\}_{k\geq 0}$ are sampled i.i.d. according to the stationary sampling distribution induced by $\pi_b$ and the transition kernel, and $\alpha_k=\alpha/(k+h)$ with appropriately chosen $\alpha$ and $h$. Then for any $K\geq 0$, with probability at least $1-\delta$ the following inequality holds for all $k\geq K$:
\begin{align*}
    \|Q_k-Q^*\|_\infty^2\leq\;&\frac{c_q\log(|\mathcal{S}||\mathcal{A}|)}{D_{b,\min}^3(1-\gamma)^5}\left(\frac{\log(1/\delta)}{k+h}+\frac{h+1+\log((k+1)/K^{1/2})}{k+h}\right),
\end{align*}
    where $c_q$ is a constant.
\end{theorem}

Theorem \ref{thm:Q-learning} is qualitatively similar to Theorem \ref{thm:additive} in that Q-learning achieves a $1/k$ convergence rate with a sub-Gaussian tail.

While Theorem \ref{thm:additive} is useful for studying Q-learning, it requires the noise to appear in an additive manner. In the context of RL, this limits the applicability of the results to the on-policy and/or tabular setting (i.e., without function approximation). To study RL in the off-policy and/or function approximation setting, we need to analyze SA with multiplicative noise, which is presented in full detail in Appendix \ref{ap:multi} for interested readers.

\subsection{Proof Sketch of Theorem \ref{thm:additive}}\label{subsec:sketch:additive}

Recall that in the establishment of classical concentration inequalities such as Hoeffding's inequality, the main idea is to bound the moment-generating function (MGF) of the random variable and then use Markov's inequality to derive the concentration results. Inspired by this approach, we develop a Lyapunov method to prove Theorem \ref{thm:additive}, where we first bound the MGF of the generalized Moreau envelope and then use Ville's maximal inequality to establish the maximal concentration bounds.

\subsubsection{Bounding the Log-MGF of the Generalized Moreau Envelope}
Let $\lambda_k = \theta / \alpha_k$, where $\alpha_k$ is the stepsize and $\theta > 0$ is a tunable parameter. 
For any $k \ge 0$, let
\begin{align*}
    Z_k = \log\left(\mathbb{E}\left[\exp\left(\lambda_k M(x_k - x^*)\right)\right]\right)
\end{align*}
be the log-MGF, where $M(\cdot)$ is the generalized Moreau envelope introduced in Section~\ref{sec:sa_mean-square} as a \textit{smooth approximation} of the norm-square function $f(x) = \frac{1}{2}\|x\|_c^2$. The properties of $M(\cdot)$ were summarized in Proposition~\ref{prop:Moreau}. We view $Z_k$ as our Lyapunov function, and the key step is to derive a bound for $Z_k$.

Working with $\mathbb{E}[\exp(\lambda_k M(x_k - x^*))]$ presents new challenges, as the nonlinear nature of the log-MGF prevents us from exploiting the linearity of expectation, which was used extensively in deriving mean-square bounds. Instead, after expressing $Z_{k+1}$ in terms of $Z_k$, we obtain the expectation of a product of random variables. To overcome this difficulty, we use a conditioning argument along with Assumption~\ref{ass:sub-Gaussian} to obtain the following inequality:
\begin{align}
    Z_k \le\;& W_1\left(\frac{h}{k+h}\right)^{\alpha D_0/2 - 1} + W_2, \quad \forall\, k \ge 0, \label{sketch:overall}
\end{align}
where $D_0$, $W_1$, and $W_2$ are problem-dependent constants.

\subsubsection{An Exponential Supermartingale and Ville's Maximal Inequality}
Let $\overline{M}_k = \exp(\lambda_k M(x_k - x^*) - W_3 \sum_{i=0}^{k-1} \alpha_i)$ for all $k \ge 0$,
where $W_3$ is a constant. Using Inequality (\ref{sketch:overall}), we show that $\{\overline{M}_k\}$ is a supermartingale with respect to the filtration $\{\mathcal{F}_k\}$ \cite{durrett2019probability}, which enables us to use Ville's maximal inequality \cite{durrett2019probability} together with Inequality~\eqref{sketch:overall} to establish a maximal concentration inequality. Specifically, for any $K \ge 0$, we have
\begin{align*}
	&\mathbb{P}\left(\sup_{k \ge K} \left\{\lambda_k M(x_k - x^*) - W_3 \sum_{i=0}^{k-1} \alpha_i\right\} > \epsilon \right)\\
	= \;&\mathbb{P}\left(\sup_{k \ge K} \left\{\exp\left(\lambda_k M(x_k - x^*) - W_3 \sum_{i=0}^{k-1} \alpha_i\right)\right\} > e^\epsilon \right)\\
	\leq \;&\exp\left(W_1\left(\frac{h}{K+h}\right)^{\alpha D_0/2 - 1} + W_2 - W_3 \sum_{i=0}^{K-1} \alpha_i - \epsilon\right),
\end{align*}
where the last line follows from Ville's maximal inequality and Inequality~\eqref{sketch:overall}. The result follows by rearranging terms in the previous inequality and then connecting $M(\cdot)$ with $\|\cdot\|_c^2$ using Proposition~\ref{prop:Moreau} (2).

\subsection{Related Work}
There is now a substantial literature on high-probability bounds for SA-type 
algorithms. For stochastic optimization, such results have been developed for 
SGD and mirror-descent-type methods under various assumptions on convexity, 
boundedness, and tail behavior of the noise 
\citep{rakhlin2012,Hazan14,Harvey19,duchi2012ergodic,heavySGD,telgarsky22}. 
For general nonlinear SA, one line of work uses the ODE method and Alekseev's 
formula to obtain lock-in or concentration bounds near stable equilibria 
\citep{thoppe2019concentration,borkar2021concentration}. Another line focuses 
on contractive SA and derives maximal concentration bounds for martingale-
difference or Markovian noise, with applications to asynchronous Q-learning and 
TD-learning 
\citep{chandak2021concentration,chen2023concentration,qian2024almost}. 
More recent work has also studied time-uniform concentration for general 
iterative algorithms and high-probability bounds for SA with Markovian and 
heavy-tailed noise \citep{pham2025timeuniform,agrawal2026concentration}.

For linear SA, high-probability bounds have been obtained for both last iterates 
and Polyak--Ruppert averaged iterates, including settings with i.i.d. and 
Markovian data 
\citep{dalal2018finite,durmus2021tight,durmus2022finite,mou2020linear}. 
In reinforcement learning, high-probability analyses have been developed for 
Q-learning and its asynchronous variants 
\citep{even2003learning,qu2020finite,li2020breaking,li2021tightening,li2024minimax}, 
as well as for TD-learning and policy evaluation with linear function 
approximation 
\citep{dalal2018finite,thoppe2019concentration,prashanth2021concentration,
patil2023finite,samsonov2024improved,li2024highprobability}. These results 
differ along several important dimensions: whether the iterates are projected 
or otherwise known to remain in a compact set, whether the noise is additive or 
multiplicative, whether the noise is bounded, sub-Gaussian, Markovian, or 
heavy-tailed, whether the stepsize depends on the target confidence level or 
time horizon, and whether the guarantee is fixed-time, finite-horizon uniform, 
or anytime. See \cite{chen2023concentration} for a more detailed comparison of 
these assumptions and guarantee formats.

\section{Future Research Directions}\label{sec:future}
There are many potential future directions for this line of work. We highlight 
three of them below.

\subsection{Multiple Time-Scale Stochastic Approximation}

The SA algorithm considered in earlier sections maintains a single sequence of 
iterates $\{x_k\}$, updated with stepsizes $\{\alpha_k\}$. In many applications, 
however, the algorithm must update several coupled sequences on different time 
scales. A canonical example is actor--critic learning in RL, where the critic 
estimates the value function on a faster time scale while the actor updates the 
policy on a slower time scale. Similar multi-time-scale structures also arise in 
gradient TD methods with auxiliary variables, constrained RL, bilevel 
optimization, and multi-agent learning.

The asymptotic theory of multiple time-scale SA has a long history and is 
typically based on the ODE method; see \cite{borkar2009stochastic} and the 
references therein. More recently, general stability and convergence conditions 
have also been developed for $N$-time-scale stochastic recursions 
\citep{deb2025multi}. In comparison, finite-time theory is more fragmented. 
Existing results cover several important special cases, including linear 
two-time-scale SA, two-timescale TD-learning, and actor--critic-type algorithms 
\citep{konda2004convergence,dalal2018finite,kaledin2020finite,doan2021finite,
doan2022nonlinear}. Recent work has further extended the Lyapunov/Moreau-envelope 
approach to nonlinear two-time-scale SA with arbitrary-norm contractions and 
Markovian noise \citep{chandak2025finite}, while another line develops accelerated 
multi-time-scale schemes that achieve $\widetilde{\mathcal O}(1/k)$ rates under 
strong monotonicity-type assumptions \citep{zeng2024accelerated}. 

Despite the progress, it remains unclear whether the obtained convergence bounds are tight in terms of their dependence on rates and other important problem-dependent parameters. For example, despite the extensive existing analyses of actor-critic methods, there are no results showing that two-timescale actor-critic can achieve the statistical lower bound \citep{gheshlaghi2013minimax}. As another example, consider the simple 
two-time-scale recursion
\begin{align}
    x_{k+1}=\;&x_k+\alpha_k(F(x_k,Y_k)-x_k),\label{sa:ttc1}\\
    \bar{x}_{k+1}=\;&\bar{x}_k+\frac{1}{k+1}(-\bar{x}_k+x_k),\label{sa:ttc2}
\end{align}
where \eqref{sa:ttc2} computes the Polyak average 
$\bar{x}_k=k^{-1}\sum_{i=0}^{k-1}x_i$. Polyak averaging is known to have strong 
asymptotic efficiency and robustness properties \citep{polyak1992acceleration}; 
for example, it can achieve the optimal asymptotic $\mathcal O(1/k)$ mean-square 
rate under stepsizes such as $\alpha_k=\alpha/(k+h)^z$ with $z\in(0,1)$. 
However, finite-time explanations of this robustness phenomenon were obtained 
only relatively recently in linear SA 
\citep{lakshminarayanan2018linear,mou2020linear,durmus2022finite}.

In general, 
a sharp and unified finite-time theory for general 
multiple time-scale SA remains incomplete. 

\subsection{Stochastic Approximation with Rapidly Time-Inhomogeneous Markovian Noise}

The Markovian-noise theory discussed in Section~\ref{subsec:extension_Markov}
assumes that the driving process $\{Y_k\}$ is a time-homogeneous Markov chain
with a fixed transition matrix $P$. In many adaptive algorithms, however, the
sampling process is itself affected by the current iterate or by another evolving
decision variable. A generic model takes the form
\begin{align*}
    x_{k+1}=x_k+\alpha_k\big(F(x_k,Y_k)-x_k\big),\qquad
    \mathbb{P}(Y_{k+1}=y'\mid \mathcal{F}_k,Y_k=y)=P_{\theta_k}(y,y'),
\end{align*}
where $\theta_k$ may be a function of $x_k$, e.g., a policy in RL, a strategy profile, or a
belief vector in learning in games. Thus, the noise process is Markovian only conditionally on the
current value of $\theta_k$, and the transition kernel changes over time.

When the map $\theta\mapsto P_\theta$ is Lipschitz and $\theta_k$ evolves
slowly, existing controlled-Markov-noise and Poisson-equation techniques can
often be adapted. Indeed, the Lipschitz dependence of $P_\theta$, the invariant
distribution, and the corresponding Poisson-equation solution allows one to
control the error caused by replacing $P_{\theta_k}$ with an earlier kernel
$P_{\theta_{k-t}}$. This is the slowly time-inhomogeneous regime, and it is
closely related to classical SA with parameter-dependent Markov chains
\citep{benveniste2012adaptive,borkar2009stochastic,care2026stochastic}.

A substantially more difficult regime arises when the transition kernel changes
rapidly with the iterate. In this case, $\|P_{\theta_{k+1}}-P_{\theta_k}\|$ need
not be of order $\|x_{k+1}-x_k\|$, and hence need not be of order $\alpha_k$.
The usual mixing-time or Poisson-equation residual terms can then contain
order-one jumps, rather than small perturbations accumulated over a mixing
window. This phenomenon is not specific to reinforcement learning. It is a
general SA issue that appears whenever the sampling kernel is generated by a
nonsmooth, discontinuous, or set-valued decision rule.

Two motivating examples illustrate the difficulty. In RL with
$\epsilon$-greedy exploration, the behavior policy is a discontinuous function
of the current $Q$-estimate: a small perturbation of $Q$ can change the identity
of the greedy action and hence produce an order-one change in the state-action
transition matrix. This discontinuity is closely related to the chattering and
policy-oscillation phenomena observed in value-based RL with $\epsilon$-greedy
exploration \citep{gopalan2023demystifying}. A similar issue arises in learning
in games. In fictitious play and best-response-type dynamics, strategies may
switch abruptly when a belief or payoff estimate crosses an indifference surface.
When the play of the agents determines the sampling distribution or the state
transition law, such strategy switches induce rapidly time-inhomogeneous
Markovian noise in the underlying SA recursion. This issue is especially natural
in stochastic games, where fictitious-play-type algorithms combine belief
updates, value-function estimates, and greedy or best-response decisions
\citep{mertikopoulos2024unified,sayin2022fictitious,baudin2022fictitious}.

Recent work has started to address special cases of this problem. For example,
\citep{nanda2025minimal} develops a finite-time analysis of Q-learning with
time-varying learning policies using a Poisson-equation decomposition and
sensitivity bounds, while \citep{liu2025linearq} studies SA with fast-changing
transition functions in the analysis of linear $Q$-learning with
$\epsilon$-softmax exploration. High-probability bounds for time-inhomogeneous
Markovian SA have also begun to appear in the analysis of online Q-learning
\citep{singh2026regret}. Nevertheless, these results do not yet constitute a
general Lyapunov theory for SA with rapidly time-inhomogeneous Markovian
noise. Developing such a theory, especially for discontinuous or set-valued
kernel maps such as $\epsilon$-greedy policies and best-response dynamics,
remains an important open direction.
\subsection{Stochastic Approximation under Nonexpansive Operators}

Another natural direction is to go beyond contractive or seminorm-contractive 
operators and consider SA driven by nonexpansive operators. An operator 
$\bar F(\cdot)$ is nonexpansive with respect to a norm $\|\cdot\|_c$ if
\begin{align*}
    \|\bar F(x_1)-\bar F(x_2)\|_c\leq \|x_1-x_2\|_c,
    \qquad \forall x_1,x_2\in\mathbb{R}^d .
\end{align*}
This setting is substantially more delicate than the contractive case. Fixed 
points need not exist; even when they exist, they need not be unique; and the 
plain fixed-point iteration $x_{k+1}=\bar F(x_k)$ need not converge to a fixed 
point. For example, $\bar F(x)=x+1$ has no fixed point, $\bar F(x)=x$ has every 
point as a fixed point, and a reflection map may cycle rather than converge.

Suppose that the fixed-point set $\mathcal X=\{x:\bar F(x)=x\}$ is nonempty. 
In the Euclidean setting, a standard way to restore convergence is to use the 
averaged operator
\begin{align*}
    \bar F_\eta(x)=(1-\eta)x+\eta \bar F(x),\qquad \eta\in(0,1),
\end{align*}
which has the same fixed-point set as $\bar F$. The resulting Krasnosel'skii--Mann 
iteration converges to a fixed point under suitable assumptions, and deterministic 
rates are typically stated in terms of the fixed-point residual, e.g.,
\begin{align*}
    \min_{0\leq i\leq k}\|\bar F_\eta(x_i)-x_i\|_2^2=\mathcal O(1/k).
\end{align*}
Finite-sample analyses of stochastic fixed-point iterations for nonexpansive 
maps have been developed for martingale-difference noise and arbitrary finite 
dimensional normed spaces \citep{bravo2024stochastic}. More recently, 
\citep{blaser2026asymptotic} studied nonexpansive SA with Markovian noise, using 
Poisson-equation decompositions to obtain both asymptotic and finite-sample 
results, with applications to tabular average-reward TD-learning.

Despite this progress, the convergence rates obtained in \citep{bravo2024stochastic,blaser2026asymptotic} imply a sample complexity of $\tilde{\mathcal{O}}(\epsilon^{-10})$ in terms of the expected residual. An important future direction is to determine whether this sample complexity can be improved, identify the corresponding fundamental lower bounds, and develop sharp, possibly optimal finite-time rates under general norms, Markovian or controlled Markovian noise, multiplicative or unbounded noise, and high-probability criteria.

\section{Conclusion}\label{sec:conclusion}
In this tutotial paper, we surveyed Lyapunov-based techniques for the finite-time analysis of stochastic approximation algorithms under contractive operators. We explained how generalized Moreau envelopes yield smooth Lyapunov functions for arbitrary contraction norms, and how the resulting drift arguments lead to mean-square and high-probability guarantees. We also discussed extensions to Markovian noise, seminorm-contractive operators, and dissipative operators, with applications to reinforcement learning algorithms such as Q-learning and TD-learning. We conclude by highlighting several open directions.

\bibliography{Arxiv/References}
\bibliographystyle{apalike}

\appendix

\newpage

\begin{center}
    \LARGE \textbf{Appendices}
\end{center}

\section{Proof of All Technical Results in Section \ref{sec:sa_mean-square}}
\subsection{Proof of Proposition \ref{prop:Moreau}}\label{pf:prop:Moreau}
\begin{enumerate}[(1)]
\item It is clear from the definition of $M_f^{\theta,g}(\cdot)$ that it is non-negative and is equal to zero if and only if $x=0$. Now for any $c\neq 0$ (the case $c=0$ is immediate), we have
\begin{align*}
	M_f^{\theta,g}(c x)&=\min_{u}\left\{\frac{1}{2}\|u\|_c^2+\frac{1}{2\theta}\|c x-u\|_s^2\right\}\\
	&= \min_{v}\left\{\frac{1}{2}\|c v\|_c^2+\frac{1}{2\theta}\|c x-c v\|_s^2\right\}\tag{change of variable $u=c v$}\\
	&=|c|^2M_f^{\theta,g}(x).
\end{align*}
Therefore, we have $\sqrt{M_f^{\theta,g}(c x)}=|c|\sqrt{M_f^{\theta,g}(x)}$.
We next show the triangle inequality. For any $x_1,x_2\in\mathbb{R}^d$, let 
\begin{align*}
    u_1\in\,&\arg\min_{u\in\mathbb{R}^d}\left\{\frac{1}{2}\|u\|_c^2+\frac{1}{2\theta}\|x_1-u\|_s^2\right\},\\
    u_2\in\,&\arg\min_{u\in\mathbb{R}^d}\left\{\frac{1}{2}\|u\|_c^2+\frac{1}{2\theta}\|x_2-u\|_s^2\right\}.
\end{align*}
Then we have
\begin{align*}
	&M_f^{\theta,g}(x_1+x_2)\\
	=\;&\min_{u}\left\{\frac{1}{2}\|u\|_c^2+\frac{1}{2\theta}\|x_1+x_2-u\|_s^2\right\}\\
	\leq \;& \frac{1}{2}\|u_1+u_2\|_c^2+\frac{1}{2\theta}\|x_1+x_2-u_1-u_2\|_s^2\tag{choose $u=u_1+u_2$}\\
	\leq \;&\frac{1}{2}(\|u_1\|_c+\|u_2\|_c)^2+\frac{1}{2\theta}(\|x_1-u_1\|_s+\|x_2-u_2\|_s)^2\\
	=\;&M_f^{\theta,g}(x_1)+M_f^{\theta,g}(x_2)+ \|u_1\|_c\|u_2\|_c+\frac{1}{\theta}\|x_1-u_1\|_s\|x_2-u_2\|_s\\
	\leq\;& M_f^{\theta,g}(x_1)+M_f^{\theta,g}(x_2)\\
    &+2\sqrt{\frac{1}{2}\|u_1\|_c^2+\frac{1}{2\theta}\|x_1-u_1\|_s^2}\sqrt{\frac{1}{2}\|u_2\|_c^2+\frac{1}{2\theta}\|x_2-u_2\|_s^2}\\
	=\;&M_f^{\theta,g}(x_1)+M_f^{\theta,g}(x_2)+2\sqrt{M_f^{\theta,g}(x_1)M_f^{\theta,g}(x_2)}.
\end{align*}
It follows that $\sqrt{M_f^{\theta,g}(x_1+x_2)}\leq \sqrt{M_f^{\theta,g}(x_1)}+\sqrt{M_f^{\theta,g}(x_2)}$ for any $x_1,x_2\in\mathbb{R}^d$.
Therefore, $M_f^{\theta,g}(\cdot)$ is a norm-square function and we can write $M_f^{\theta,g}(x)=\frac{1}{2}\|x\|_m^2$ for some norm $\|\cdot\|_m$.
\item We first derive the upper bound. By definition of $	M_f^{\theta,g}(\cdot)$, we have
\begin{align*}
	M_f^{\theta,g}(x)&=\min_{u\in\mathbb{R}^d}\left\{\frac{1}{2}\|u\|_c^2+\frac{1}{2\theta}\|x-u\|_s^2\right\}\\
	&\geq \min_{u\in\mathbb{R}^d}\left\{\frac{1}{2}\|u\|_c^2+\frac{1}{2\theta u_{cs}^2}\|x-u\|_c^2\right\}\tag{$\|\cdot\|_c\leq u_{cs}\|\cdot\|_s$}\\
	&\geq \min_{u\in\mathbb{R}^d}\left\{\frac{1}{2}\|u\|_c^2+\frac{1}{2\theta u_{cs}^2}(\|x\|_c-\|u\|_c)^2\right\}\tag{triangle inequality}\\
	&=\min_{y\in\mathbb{R}}\left\{\frac{1}{2}y^2+\frac{1}{2\theta u_{cs}^2}(\|x\|_c-y)^2\right\}\tag{change of variable: $y=\|u\|_c$}\\
	&=\min_{y\in\mathbb{R}}\left\{\left(\frac{1}{2}+\frac{1}{2\theta u_{cs}^2}\right)y^2-\frac{1}{\theta u_{cs}^2}\|x\|_cy+\frac{1}{2\theta u_{cs}^2}\|x\|_c^2\right\}\\
	&=\frac{1}{2}\|x\|_c^2\frac{1}{\theta u_{cs}^2+1}\tag{minimum of a quadratic function}\\
	&=\frac{1}{\theta u_{cs}^2+1}f(x).
\end{align*}
It follows that $f(x)\leq \left(1+\theta u_{cs}^2\right) M_f^{\theta,g}(x)$ for all $x$, which implies $\|\cdot\|_c\leq (1+\theta u_{cs}^2)^{1/2}\|\cdot\|_m$.

Next we show the lower bound. Similarly, by definition we have for any $x\in\mathbb{R}^d$ that
\begin{align*}
	M_f^{\theta,g}(x)&=\min_{u\in\mathbb{R}^d}\left\{\frac{1}{2}\|u\|_c^2+\frac{1}{2\theta}\|x-u\|_s^2\right\}\\
	&\leq \min_{\alpha\in(0,1)}\left\{\frac{1}{2}\|\alpha x\|_c^2+\frac{1}{2\theta}\|x-\alpha x\|_s^2\right\}\tag{restrict $u=\alpha x$ for $\alpha\in (0,1)$}\\
	&\leq \frac{1}{2}\|x\|_c^2\min_{\alpha\in(0,1)}\left\{\alpha^2+\frac{(1-\alpha)^2}{\theta \ell_{cs}^2}\right\}\tag{$\ell_{cs}\|\cdot\|_s\leq \|\cdot\|_c$}\\
	&=\frac{1}{1+\theta\ell_{cs}^2}\frac{1}{2}\|x\|_c^2\tag{minimum of the quadratic function}\\
	&= \frac{1}{1+\theta\ell_{cs}^2}f(x).
\end{align*}	
It follows that $f(x)\geq \left(1+\theta \ell_{cs}^2\right)M_f^{\theta,g}(x)$ for all $x$, which implies $\|\cdot\|_c\geq (1+\theta \ell_{cs}^2)^{1/2}\|\cdot\|_m$.
\item The convexity of $M_f^{\theta,g}(\cdot)$ follows from  \cite[Theorem 2.19]{beck2017first}.
Since $f(\cdot)$ is proper, closed, and convex, and $g(\cdot)$ is $L$ -- smooth with respect to $\|\cdot\|_s$, we have by \cite[Theorem 5.30 (a)]{beck2017first}  that $M_f^{\theta,g}(\cdot)$ is $\frac{L}{\theta}$ -- smooth with respect to $\|\cdot\|_s$.
\end{enumerate}

\subsection{Proof of Lemma \ref{le:T1}}\label{pf:le:T1}
Using the fact that $\bar{F}(x^*)=x^*$, we have
\begin{align}\label{eq:split-E-terms}
	&\langle \nabla M_f^{\theta,g}(x_k-x^*),\bar{F}(x_k)-x_k\rangle\nonumber\\
	=\;&\underbrace{\langle \nabla M_f^{\theta,g}(x_k-x^*),\bar{F}(x_k)-\bar{F}(x^*)\rangle}_{E_{1,1}}-\underbrace{\langle \nabla M_f^{\theta,g}(x_k-x^*),x_k-x^*\rangle}_{E_{1,2}}.
\end{align}
For the gradient of $M_f^{\theta,g}(x)$, since $M_f^{\theta,g}(x)=\frac{1}{2}\|x\|_m^2$ is a smooth function, we have by the chain rule of calculus that $\nabla M_f^{\theta,g}(x)=\|x\|_m \nabla \|x\|_m$. 

Now consider the term $E_{1,1}$. Using H\"{o}lder's inequality, we have
\begin{align}
	E_{1,1}&=\|x_k-x^*\|_m\langle \nabla \|x_k-x^*\|_m,\bar{F}(x_k)-\bar{F}(x^*)\rangle\nonumber\\
	&\leq \|x_k-x^*\|_m\|\nabla \|x_k-x^*\|_m\|_m^*\|\bar{F}(x_k)-\bar{F}(x^*)\|_m,\label{eq:dual-norm}
\end{align}
where $\|\cdot\|_m^*$ is the dual norm of $\|\cdot\|_m$. To further control $E_{1,1}$, the following result from \cite{shalev2012online} is needed.

\begin{lemma}\label{le:Lipschitz}
	Let $\hat{h}: \mathcal{D} \rightarrow \mathbb{R}$ be a convex differentiable function. Then $\hat{h}$ is $L'$ -- Lipschitz over $\mathcal{D}$ with respect to some norm $\|\cdot\|$ if and only if $\sup_{w\in\mathcal{D}}\|\nabla \hat{h}(w)\|_{*}\leq L'$, where $\|\cdot\|_{*}$ is the dual norm of $\|\cdot\|$.
\end{lemma}

Since $\|x\|_m$ as a function of $x$ is $1$ -- Lipschitz with respect to $\|\cdot\|_m$, we have by Lemma \ref{le:Lipschitz} that $\|\nabla \|x_k-x^*\|_m\|_m^*\leq 1$. For the term $\|\bar{F}(x_k)-\bar{F}(x^*)\|_m$ in Inequality (\ref{eq:dual-norm}), using Proposition \ref{prop:Moreau} (2) and the contraction property of $\bar{F}(\cdot)$ with respect to $\|\cdot\|_c$, we have
\begin{align*}
	\|\bar{F}(x_k)-\bar{F}(x^*)\|_m\leq \,&\frac{1}{\ell_{cm}}\|\bar{F}(x_k)-\bar{F}(x^*)\|_c\\
    \leq \,&\frac{\gamma_c}{\ell_{cm}}\|x_k-x^*\|_c\\
    \leq \,&\frac{\gamma_c u_{cm}}{\ell_{cm}}\|x_k-x^*\|_m.
\end{align*}
Substituting the upper bounds we obtained for $\|\nabla \|x_k-x^*\|_m\|_m^*$ and $\|\bar{F}(x_k)-\bar{F}(x^*)\|_m$ into Inequality (\ref{eq:dual-norm}), we have
\begin{align*}
	E_{1,1}
	\leq \frac{\gamma_c u_{cm}}{\ell_{cm}}\|x_k-x^*\|_m^2=\frac{2\gamma_c u_{cm}}{\ell_{cm}} M_f^{\theta,g}(x_k-x^*).
\end{align*}

Now consider the term $E_{1,2}$ in Inequality (\ref{eq:split-E-terms}). Since the norm $\|\cdot\|_m$ is a convex function of $x$, we have by definition of convexity that
\begin{align*}
    \|0\|_m-\|x_k-x^*\|_m\geq \langle \nabla \|x_k-x^*\|_m,-(x_k-x^*)\rangle,
\end{align*}
which implies
\begin{align*}
	E_{1,2}=\|x_k-x^*\|_m\langle \nabla \|x_k-x^*\|_m,x_k-x^*\rangle\geq \|x_k-x^*\|_m^2=2M_f^{\theta,g}(x_k-x^*).
\end{align*}
Combining the bounds on $E_{1,1}$ and $E_{1,2}$, we obtain
\begin{align*}
	E_1=\alpha_k (E_{1,1}-E_{1,2})\leq -2\left(1-\gamma_c\frac{u_{cm}}{\ell_{cm}}\right)\alpha_k M_f^{\theta,g}(x_k-x^*).
\end{align*}

\subsection{Proof of Lemma \ref{le:T3}}\label{pf:le:T3}
We first show that $\|F(x,y)\|_c$ can grow at most affinely with respect to $\|x\|_c$. Indeed, by the triangle inequality, we have
\begin{align}\label{eq:affine_growth}
    \|F(x,y)\|_c 
    \leq  \|F(x,y) - F(0,y)\|_c + \|F(0,y)\|_c \leq L_1 \|x\|_c + L_2.
\end{align}
Further, by Jensen's inequality, we have
\begin{align*}
    \|\bar{F}(x)\|_c\leq L_1\|x\|_c+L_2.
\end{align*}
To bound the term $E_3$, for any $k\geq 0$, we have
\begin{align*}
	E_3&=\frac{L\alpha_k^2}{2\theta}\|F(x_k,Y_k)-x_k\|_s^2\\
	&\leq \frac{L\alpha_k^2}{2\theta\ell_{cs}^2}\|F(x_k,Y_k)-x_k\|_c^2\tag{Proposition \ref{prop:Moreau} (2)}\\
	&\leq \frac{L\alpha_k^2}{2\theta\ell_{cs}^2}\left(\|F(x_k,Y_k)\|_c+\|x_k\|_c\right)^2\\
	&\leq \frac{L\alpha_k^2}{2\theta\ell_{cs}^2}\left((L_1+1)\|x_k\|_c+L_2\right)^2\tag{Inequality~(\ref{eq:affine_growth})}\\
	&\leq \frac{L\alpha_k^2}{2\theta\ell_{cs}^2}\left((L_1+1)\|x_k-x^*\|_c+(L_1+1)\|x^*\|_c+L_2\right)^2\\
	&\leq \frac{L\alpha_k^2}{\theta\ell_{cs}^2}(L_1+1)^2\|x_k-x^*\|_c^2+\frac{L\alpha_k^2}{\theta\ell_{cs}^2}\left((L_1+1)\|x^*\|_c+L_2\right)^2\\
	&\leq \frac{2L(L_1+1)^2u_{cm}^2\alpha_k^2}{\theta\ell_{cs}^2}M_f^{\theta,g}(x_k-x^*)+\frac{L\alpha_k^2}{\theta\ell_{cs}^2}\left((L_1+1)\|x^*\|_c+L_2\right)^2,
\end{align*}
where the last line follows from Proposition~\ref{prop:Moreau}~(2).

\subsection{Proof of Lemma \ref{le:MSE_drift}}\label{pf:le:MSE_drift}
Using Lemma~\ref{le:T1}, Equation~(\ref{eq:T_2}), and Lemma~\ref{le:T3} in Inequality~(\ref{eq:composition1_contractive}), we have for all $k\geq 0$ that
\begin{align*}
    M_{k+1}\leq\,& \left(1- 2\left(1 - \frac{\gamma_c u_{cm}}{\ell_{cm}}\right)\alpha_k+\frac{2L(L_1+1)^2u_{cm}^2\alpha_k^2}{\theta\ell_{cs}^2}\right)M_k\\
    &+\frac{L\alpha_k^2}{\theta\ell_{cs}^2}\left((L_1+1)\|x^*\|_c+L_2\right)^2\\
    \leq \,&\left(1-\left(1 - \frac{\gamma_c u_{cm}}{\ell_{cm}}\right)\alpha_k \right)M_k+\frac{L\alpha_k^2}{\theta\ell_{cs}^2}\left((L_1+1)\|x^*\|_c+L_2\right)^2,
\end{align*}
where the last inequality follows from choosing
\begin{align*}
    \alpha_0\leq \varphi_0=\frac{\theta \ell_{cs}^2\left(1 - \gamma_c u_{cm}/\ell_{cm}\right)}{2 L (L_1+1)^2 u_{cm}^2 }.
\end{align*}

\section{Stochastic Approximation under Bounded Multiplicative Noise}\label{ap:multi}

The following assumption explains what we mean by multiplicative noise.

\begin{assumption}\label{as:multi}
	There exists $\sigma>0$ such that $\| F(x_k,Y_k) - \Bar{F}(x_k) \|_c \leq \sigma(1+\|x_k\|_c)$ a.s. for all $k\geq 0$.
\end{assumption}

Assumption \ref{as:multi} is in fact a relaxed version of Assumption \ref{as:Lipschitz}. To see this, suppose that Assumption \ref{as:Lipschitz} is satisfied. Then we have the affine growth property (cf. Equation (\ref{eq:affine_growth})), which in turn implies that 
	\begin{align*}
		\| F(x_k,Y_k) - \Bar{F}(x_k) \|_c\leq \;&\| F(x_k,Y_k)\|_c +\| \Bar{F}(x_k) \|_c\\
		= \;&\| F(x_k,Y_k)\|_c +\| \mathbb{E}_{Y\sim \mu}[F(x_k,Y)] \|_c\\
		\leq \;&\| F(x_k,Y_k)\|_c +\mathbb{E}_{Y\sim \mu}[\| F(x_k,Y) \|_c]\tag{Jensen's inequality}\\
		\leq\;& 2L_1\|x_k\|_c+2L_2.
	\end{align*}
	Therefore, Assumption \ref{as:multi} is satisfied with $\sigma=2\max(L_1,L_2)$.

The almost sure boundedness in Assumption \ref{as:multi} is necessary for us to establish high probability bounds with super-polynomial tail. To see this, consider scalar-valued linear SA $x_{k+1}=x_k+\alpha_k\tilde{a}_kx_k$, and suppose that $\tilde{a}_k\sim \mathcal{N}(-1/2,1)$. Note that $\tilde{a}_k$ is negative in expectation but is not bounded. It is easy to see that $x_k=x_0\prod_{j=0}^{k-1}(1+\alpha_j \tilde{a}_j)$.
Since $1+\alpha_k \tilde{a}_k\sim \mathcal{N}(1-\alpha_k/2,\alpha_k^2)$, the tail of $x_k$ gets heavier and heavier as $k$ increases. For example, $x_1$ has sub-Gaussian tail, $x_2$ has a $\mathcal{X}^2$-distribution (which only has sub-exponential tail), $x_3$ has no MGF, etc. One special case where Assumption \ref{as:multi} is satisfied is when the operator $F(x,y)$ is Lipschitz continuous in $x$ (cf. Assumption \ref{as:Lipschitz}). The implication easily follows from using triangle inequality. 

Observe that under Assumptions \ref{as:contraction} and \ref{as:multi}, we have by triangle  inequality that
\begin{align*}
	\|x_{k+1}\|_c\leq\;& (1-\alpha_k)\|x_k\|_c+\alpha_k\|F(x_k,Y_k)\|_c\\
	\leq \;&(1-\alpha_k)\|x_k\|_c+\alpha_k\|F(x_k,Y_k)-\bar{F}(x_k)\|_c\\
    &+\alpha_k\|\bar{F}(x_k)-\bar{F}(0)\|_c+\alpha_k\|\bar{F}(0)\|_c\\
	\leq \;&(1+(\sigma+\gamma_c-1)\alpha_k)\|x_k\|_c+\alpha_k (\sigma+\|\bar{F}(0)\|_c).
\end{align*}
Note that the parameter $\gamma_c$ captures the contraction effect of the expected operator and the parameter $\sigma$ captures the expansive effect of the noise. The combined effect is captured by the parameter $D:=\sigma+\gamma_c-1$. When using $\alpha_k=\alpha/(k+h)$, the previous inequality implies that $\|x_k\|_c$ is either uniformly bounded by a deterministic constant or grow at most logarithmically when $D\leq 0$. However, when $D>0$, $\|x_k\|_c$ can grow at a polynomial rate $\mathcal{O}(k^{\alpha D})$. See Proposition \ref{prop:worst_case_bound} for more details.

We next state the result. For ease of exposition, we only state the result for $D>0$ and $D<0$. The case $D=0$ is a straightforward extension. We use $\alpha_k=\alpha/(k+h)$ as our stepsize, and assume that $2\alpha D$ is an integer, which is in fact without loss of generality because $D=\sigma+\gamma_c-1$ and if Assumption \ref{as:Lipschitz} holds with some $\sigma>0$ it also holds for any $\sigma'>\sigma$. Let $m=2\alpha D+1$. The parameters $\{c_i\}_{1\leq i\leq 4}$, $c_1'$, and $D_0\in (0,1)$ used to present the following theorem are constants.
\begin{theorem}[\cite{chen2023concentration}]\label{thm:multi}
	Consider $\{x_k\}$ generated by Algorithm (\ref{algo:sa1}). Suppose that Assumptions \ref{as:iid}, \ref{as:contraction}, and \ref{as:multi} are satisfied. 
	\begin{enumerate}[(1)]
		\item When $D<0$, $\alpha>2/D_0$, and $h$ is appropriately chosen, for any $\delta>0$ and $K\geq 0$, with probability at least $1-\delta$, we have for all $k\geq K$ that
		\begin{align*}
			\|x_k-x^*\|_c^2
			\leq \;&\frac{c_1'\alpha\|x_0-x^*\|_c^2}{k+h}\left[\log\left(\frac{1}{\delta}\right)+c_2\left(\frac{h}{K+h}\right)^{\alpha D_0/2-1}\right.\\
            &\left.\,+c_3+c_4 \log\left(\frac{k-1+h}{K-1+h}\right)\right].
		\end{align*}
		\item When $D>0$, $\alpha>2/D_0$, and $h$ is appropriately chosen, for any $\delta>0$ and $K\geq 0$, with probability at least $1-\delta$, we have for all $k\geq K$ that
		\begin{align*}
			&\|x_k-x^*\|_c^2
			\leq \frac{c_1\alpha \|x_0\!-\!x^*\|_c^2}{k+h} \left[\log\left(\frac{m}{\delta}\right)\!+\!c_2\!+\!c_3\!+\!c_4\log\left(\frac{k-1+h}{h-1}\right)\right]^{m-1}\nonumber\\
			&\quad \times \left[\log\left(\frac{m}{\delta}\right)+c_2\left(\frac{h}{K+h}\right)^{\alpha D_0/2-1}+c_3+c_4\log\left(\frac{k-1+h}{K-1+h}\right)\right].
		\end{align*}
	\end{enumerate}
\end{theorem}

Several remarks are in order. We first discuss the convergence rate in $k$. Since the dominant polynomial term in $k$ is $\mathcal{O}(1/k)$ in both cases of Theorem \ref{thm:multi} (i.e., $D<0$ and $D>0$), the norm-square error achieves $\mathcal{O}(1/k)$ rate of convergence. This agrees with the mean-square bound presented in Section \ref{sec:sa_mean-square}, and existing results on SGD \cite{lan2020first}.

We next discuss the tail. In the case $D<0$, Theorem \ref{thm:multi} (1) states that Algorithm (\ref{algo:sa1}) achieves $\mathcal{O}(1/k)$ rate of convergence with sub-Gaussian tail, as $\delta$ appears as $\log(1/\delta)$ in the norm-square bound. This is not surprising as $D<0$ implies that $\|x_k\|_c$ is uniformly bounded by a deterministic constant (cf. Proposition \ref{prop:worst_case_bound}). Therefore, we in fact only have bounded (hence sub-Gaussian) additive noise.

The case where $D>0$ is more complicated. In this case, the tail depends on the parameter $m$. Since $m=2\alpha D+1$ and $D>0$, we in general only have super-polynomial tail since the probability tolerance level $\delta$ appears as $[\log(m/\delta)]^m$ in the bound. The fact that $m$ is affine in $D$ makes intuitive sense as larger $D$ implies noisier updates, which in turn implies heavier tail. In terms of the dependency on $k$ and $K$,
Theorem \ref{thm:multi} (2) states that, with probability at least $1-\delta$, all the iterates lie in a cone that starts with radius $\tilde{\Theta}((1+\log^{m/2}(1/\delta))K^{-1/2})$ when $k=K \geq 0$, and then (for all $k>K$) its radius is of order $\tilde{\Theta}((\log^{m/2}(1/\delta)+\log^{1/2}(k/K))k^{-1/2})$. Moreover, for small values of $k$, this bound can be tightened by an a.s. bound that is polynomial in $k$ (cf. Proposition \ref{prop:worst_case_bound}). Note that, as a function of $k$, the initial radius is always of order at most $K^{-1}$, matching the rate obtained for the mean-square error in \cite{chen2024lyapunov}. On the other hand, the radius decays at only a slightly slower rate than the initial radius as a function of $k$.

Anytime concentration bounds immediately imply concentration bounds for a fixed iteration number (cf. Corollary \ref{cor:bounded}), which in turn gives the full tail bound (cf. Corollary \ref{cor:boundedTail}) and the sample complexity result (cf. Corollary \ref{co:sc}). For ease of exposition, we only state the results when $D>0$.

\begin{corollary} \label{cor:bounded}
	Suppose that $D>0$. Under the same assumptions in Theorem \ref{thm:multi} (2), for any $\delta>0$ and $k\geq 0$, we have with probability at least $1-\delta$ that
	\begin{align*}
		\|x_k-x^*\|_c^2
		\leq \;&\frac{c_1\alpha \|x_0-x^*\|_c^2}{k+h} \left[\log\left(\frac{m}{\delta}\right)+c_2+c_3+c_4\log\left(\frac{k-1+h}{h-1}\right)\right]^{m}.
	\end{align*}
\end{corollary}

Corollary \ref{cor:bounded} follows by setting $K=k$ in Theorem \ref{thm:multi} (2).

\begin{corollary}\label{cor:boundedTail}
	Suppose that $D>0$. Under the same assumptions in Theorem \ref{thm:multi} (2), there exists $C_1>0$ such that the following inequality holds for all $\epsilon>0$ and $k\geq 0$: 
	\begin{align*}
		\mathbb{P}\left(\frac{\sqrt{k+h}\;\| x_k - x^* \|_c}{(\log(k))^{m/2}}> \epsilon \right) < m\exp\left(-C_1\epsilon^{2/m} \right).
	\end{align*}
\end{corollary}

Corollary \ref{cor:boundedTail} is a direct implication of Corollary \ref{cor:bounded}, and provides an upper bound for the whole complementary cumulative distribution function (CDF) of the error $\|x_k-x^*\|_c$ for any iterate $k\geq 0$, which can be integrated to obtain bounds for any moment of the error at any point in time.

\begin{corollary}\label{co:sc}
	Given $\epsilon>0$, to achieve $\|x_k-x^*\|_c\leq \epsilon$ with probability at least $1-\delta$, the sample complexity is $\tilde{\mathcal{O}}((1+\log^{m}(1/\delta))\epsilon^{-2})$.
\end{corollary}

As we see from Corollary \ref{co:sc}, the sample complexity dependency on $\epsilon$ is $\tilde{\mathcal{O}}(\epsilon^{-2})$, which is known to be optimal (up to  a logarithmic factor). In addition, we have super-polynomial tail as $\delta$ appears as $\log^m(1/\delta)$ in the bound. 

Theorem \ref{thm:multi} shows that SA with multiplicative noise in general is able to achieve an $\tilde{\mathcal{O}}(1/k)$ rate of convergence with a super-polynomial tail. One may ask if sub-Gaussian (or sub-exponential) tail is achievable. In fact, for SA with multiplicative noise, it is in general not possible to obtain sub-exponential tail bound. See \cite{chen2023concentration} for more details.

\subsection{Proof Sketch of Theorem \ref{thm:multi}}\label{subsec:sketch:multi}
The main challenge of obtaining super-polynomial high probability bounds is due to the combination of \textit{unbounded iterates} and \textit{multiplicative noise}. While having unbounded iterates and multiplicative noise are not too problematic in isolation, the combination of both creates a setting where the variance of the noise is unbounded. In this case, since we allow the multiplicative noise to be large enough so that the ``noisy'' operator can be expansive with positive probability, the error can grow extremely fast with a significant probability. This creates a challenge that no approach in the literature can deal with in general. To overcome this challenge, we develop a novel bootstrapping argument. The high level ideas are presented in the following.

\subsubsection{Initialization: Time-Varying Worst-Case Bounds} While the iterates of SA with multiplicative noise are not uniformly bounded by a constant, we show in the following proposition that they do admit a time-varying a.s. bound. The behavior of such time-varying bound depends on the contraction effect in the expected operator and the expansive effect in the multiplicative noise. In general, the bound can be polynomially \textit{increasing} with time.

\begin{proposition}[\cite{chen2023concentration}]\label{prop:worst_case_bound}
	Consider $\{x_k\}$ generated by Algorithm (\ref{algo:sa1}). Suppose that Assumptions \ref{as:contraction} and  \ref{as:multi} are satisfied, and $\alpha_k=\alpha/(k+h)$ for all $k\geq 0$, where $\alpha,h>0$ are constants. Then we have $\|x_k-x^*\|_c\leq B_k(D)$ a.s. for all $k\geq 0$, where
	\begin{align*}
		B_k(D)=\begin{dcases}
			\left(\frac{k\!-\!1\!+\!h}{h\!-\!1}\right)^{\alpha D}\left(\|x_0\!-\!x^*\|_c\!+\!\frac{\sigma(1\!+\!\|x^*\|_c)}{D}\right)\!-\!\frac{\sigma(1\!+\!\|x^*\|_c)}{D},&D> 0,\\
			\|x_0-x^*\|_c+\sigma(1+\|x^*\|_c)\alpha \log\left(\frac{k-1+h}{h-1}\right),&D=0,\\
			\|x_0-x^*\|_c-\frac{\sigma(1+\|x^*\|_c)}{D},&D< 0.
		\end{dcases} 
	\end{align*}
\end{proposition}

\subsubsection{Bootstrapping: An Iterative Framework to Improve the Bound} The key in the bootstrapping argument is to start with a non-decreasing sequence $\{T_k(\delta)\}_{k\geq 0}$ such that 
\begin{align*}
	\mathbb{P}(\|x_k-x^*\|_c^2\leq T_k(\delta),\forall\;k\geq 0)\geq 1-\delta,
\end{align*}
and obtain a sequence $\{T_k(\delta,\delta')\}_{k\geq 0}$, with $T_k(\delta,\delta')=\tilde{\mathcal{O}}(T_k(\delta)/k)$, such that 
\begin{align*}
	\mathbb{P}(\|x_i-x^*\|_c^2\leq T_k(\delta,\delta'),\forall\;k\geq 0)\geq 1-\delta-\delta'.
\end{align*}
This blueprint enables us to start with the time-varying worst-case bound for the error (which can be polynomially increasing) and iteratively improve it to obtain our super-polynomial concentration bound with the desired convergence rate. To establish this blueprint, we develop a refined variant of the two-step Lyapunov approach presented in Section \ref{subsec:sketch:additive}. 
\end{document}